\documentclass{article}

\usepackage{microtype}
\usepackage{graphicx}
\usepackage{subfigure}
\usepackage{booktabs} 
\usepackage{hyperref}
\usepackage{setspace}
\usepackage{multirow}
\usepackage{makecell}

\usepackage[accepted]{arxiv2023}

\usepackage{amsmath}
\usepackage{amssymb}
\usepackage{mathtools}
\usepackage{amsthm}

\usepackage[capitalize,noabbrev]{cleveref}

\usepackage{tikz}
\usetikzlibrary{shapes.geometric, arrows}
\usetikzlibrary{positioning}
\usetikzlibrary{arrows.meta}
\usetikzlibrary{decorations.pathreplacing,calligraphy}
\theoremstyle{plain}
\newtheorem{theorem}{Theorem}[section]
\newtheorem{proposition}[theorem]{Proposition}
\newtheorem{lemma}[theorem]{Lemma}

\theoremstyle{definition}

\theoremstyle{remark}

\usepackage[textsize=tiny]{todonotes}

\usepackage{caption}
\usepackage{subcaption}
\usepackage{enumitem}
\usepackage{wrapfig}
\usepackage{arydshln}
\newcommand{\modelaname}{\textit{OSSCAR}}
\newcommand{\Conv}{\operatorname{Conv}}
\newcommand{\Tr}{\operatorname{Tr}}
\newcommand{\wbf}{\mathbf{w}}
\newcommand{\modelanamelong}{\textbf{O}ne-\textbf{S}hot
\textbf{S}tructured \textbf{C}ompression \textbf{A}lgo\textbf{R}ithm
}

\icmltitlerunning{\modelaname: One-Shot Structured Pruning}

\begin{document}

\twocolumn[
\icmltitle{OSSCAR: One-Shot Structured Pruning in Vision and Language Models with Combinatorial Optimization}




\begin{icmlauthorlist}
\icmlauthor{Xiang Meng}{school}
\icmlauthor{Shibal Ibrahim}{school}
\icmlauthor{Kayhan Behdin}{school}
\icmlauthor{Hussein Hazimeh}{company}
\icmlauthor{Natalia Ponomareva}{company}
\icmlauthor{Rahul Mazumder}{school}
\end{icmlauthorlist}

\icmlaffiliation{school}{Massachusetts Institute of Technology, Cambridge, MA, USA}
\icmlaffiliation{company}{Google Research, New York, NY, USA}

\icmlcorrespondingauthor{Xiang Meng}{mengx@mit.edu}

\icmlkeywords{Structured Pruning, Channel Pruning, Neural Networks, Large Language Models, Convolutional Neural Networks}

\vskip 0.3in
]



\printAffiliationsAndNotice{}  

\begin{abstract}

Structured pruning is a promising approach for reducing the inference costs of large vision and language models. By removing carefully chosen structures, e.g., neurons or attention heads, the improvements from this approach can be realized on standard deep learning hardware. In this work, we focus on structured pruning in the one-shot (post-training) setting, which does not require model retraining after pruning. We propose a novel combinatorial optimization framework for this problem, based on a layer-wise reconstruction objective and a careful reformulation that allows for scalable optimization. Moreover, we design a new local combinatorial optimization algorithm, which exploits low-rank updates for efficient local search. Our framework is time and memory-efficient and considerably improves upon state-of-the-art one-shot methods on vision models (e.g., ResNet50, MobileNet) and language models (e.g., OPT-1.3B -- OPT-30B). 
For language models, e.g., OPT-2.7B, \modelaname~can lead to $125\times$ lower test perplexity on WikiText with $2\times$ inference time speedup in comparison to the state-of-the-art ZipLM approach. Our framework is also $6\times$ -- $8\times$ faster.
Notably, our work considers models with tens of billions of parameters, which is up to $100\times$ larger than what has been previously considered in the structured pruning literature.
\end{abstract}

\section{Introduction}
Structured pruning \citep{Lebedev2016,Wen2016} reduces model size by removing entire subcomponents, e.g., channels in convolutional layers, neurons in dense layers and heads in multi-head attention.
Structured pruning offers a practical solution to improve inference latency on standard hardware 
in contrast to unstructured pruning \citep{OBD,OBS1992,Han2015}, which requires specialized hardware and software.


Although there are direct computational benefits from structured pruning, there is a significant challenge.
Models can be highly sensitive to structured pruning (with a large drop in utility) and most existing methods \citep{Li2017,Molchanov2017,He2017,Luo17,Yu2018} rely on gradual pruning or iterative retraining, where the model is finetuned on the original loss after every pruning stage to allow the model to recover accuracy. 
Such finetuning may not be desirable for large datasets and models under resource constraints. For example, finetuning an LLM on a standard GPU (A100) may not be possible beyond a few billion parameters~\citep{Malladi2023}. 
In this context, recent works \citep{kwon2022a,ziplm} have focused on the challenging task of post-training one-shot structured pruning, in which a model must be compressed (without retraining) based on a small amount of calibration data, without significant loss in accuracy. 


Despite impressive advances, state-of-the-art methods appear to face challenges in terms of increased computation time, memory usage, and balancing utility with structured sparsity. 
To address these challenges, we propose a novel optimization-based framework for one-shot structured pruning, which is highly scalable and can achieve good utility-sparsity tradeoffs. 
We employ a layer-wise reconstruction objective and a careful structure-aware reformulation of the optimization formulation to enable more scalability. 
To obtain good solutions to the problem, we propose a local combinatorial optimization algorithm, which leverages problem structure to perform efficient local search.
By integrating these algorithmic components, we demonstrate that our method is capable of handling large vision and language models, up to 30 billion parameters using a single 32GB V100 GPU --- an improvement over prior approaches that can handle up to 340 million parameters.

\paragraph{Contributions.} Our technical contributions are:
\begin{enumerate}[noitemsep,topsep=0pt,parsep=0pt,partopsep=0pt, leftmargin=*]
    \item Motivated by prior work \citep{He2017,Dong2017,ziplm}, we consider a layer-wise reconstruction objective.
    We formulate the structured pruning problem as a quadratic program with combinatorial constraints. Leveraging the loss structure by identifying groups of variables sharing the same quadratic coefficients, we propose a reformulation of the objective that significantly reduces the scale of the quadratic coefficient matrix.
    Our framework, named as \modelaname\footnote{\modelaname: \modelanamelong}, is a novel general structured pruning framework that prunes channels in convolutional layers, neurons in dense layers or heads in multi-head attention layers.
    \item The combinatorial nature of structured pruning makes it a challenging optimization problem. We propose a new efficient algorithm, which performs local combinatorial search to (i) iteratively prune structures to satisfy desired combinatorial constraints, as well as (ii) locally explore structures with similar structured sparsity budgets, but with smaller objectives. 
    Our algorithm exploits novel low-rank matrix updates in its combinatorial search and enjoys theoretical guarantees (time and memory cost analysis). 
    \item \modelaname~considerably improves over state-of-the-art approaches for one-shot structured pruning of language and vision models, both in terms of achieved quality and inference time. (a) For language models, e.g., OPT-2.7B, \modelaname~can lead to $125\times$ lower test perplexity on WikiText with $2\times$ inference time speedup in comparison to state-of-the-art  
    ZipLM \citep{ziplm} approach. Our framework is also $6\times$ -- $8\times$ faster than ZipLM. 
    \modelaname~achieves $1.6\times$ storage reduction in saving decoder layers over ZipLM, when we match performance.
    (b) For vision models, e.g., ResNet50, \modelaname~achieves $10\%$ better accuracy for $\sim 2\times$ speedup over previous state-of-the-art approaches when adapted to one-shot setting. 
\end{enumerate}
Notably, we are the first to consider up to $30$ billion model sizes for structured pruning, which is $100\times$ larger than what has been considered by prior works on structured pruning \citep{kwon2022a,ziplm}. The existing state-of-the-art pipelines are unable to scale to structured pruning of model with such sizes. Our code will be open-sourced if the paper gets accepted.

\section{Related Work}
Network pruning can be generally categorized into unstructured
and structured methods.
(a) Unstructured methods \citep{OBD,OBS1992,Han2015,Han2016,Guo2016} prune unimportant weights in the model, but efficiency of the pruned sparse network cannot be realized on general-purpose GPU hardware.
(b)~Structured methods prune channels,  neurons etc. \citep{Lebedev2016,Wen2016}, and thus actual speedup can be realized without the need for sparse accelerators. In this work, we consider structured pruning. Next, we summarize structured pruning methods in vision and language models. 

\looseness=-1 \noindent\textbf{Structured pruning in Vision Models.}
\citet{Lebedev2016,Wen2016,Zhou2016} consider $\ell_{21}$ norm to prune filters in convolutional layers.
\citet{Li2017} and \citet{He2018a} use $\ell_1$ norm and $\ell_2$ norm of the filters for pruning respectively.
\citet{He2017,Luo17} minimize a layer-wise reconstruction error using LASSO and greedy methods to prune channels, respectively.
\citet{Yu2018}'s approach minimizes reconstruction error of the final response layer and propagates importance scores through the entire network.
\citet{Molchanov2017} uses first-order information from Taylor expansion to greedily prune the least important channels.
\citet{Liu2017,Luo2020} use scaling factors in Batch Normalization layers as importance scores for pruning channels.
\citet{liu21ab} use Fisher-approximation based importance scores to prune channels (or groups of channels).
\citet{Tang2020,Lin2020cvpr,Sui2021} consider feature-importance based measure to determine the important channels.
All the above methods are used in the context of gradual pruning, or finetuning, which is prohibitively expensive on large pretrained models and datasets.
We explore one-shot pruning adaptations of some of these approaches, and compare them with our approach.

\noindent\textbf{Structured Pruning in Language Models.}
Previous works focus on pruning different components in Transformer models.
\citet{Voita2019,Michel2019} prune heads in multi-head attention layers. 
\citet{McCarley2019,Hou2020,Chen2021earlybert} prune the feed-forward network (FFN) by reducing the intermediate dimension. 
\citet{Fan2020,Sajjad2020} drop entire Transformer blocks (a pair of MHA and FFN) from a pre-trained model.
All of these methods are paired with gradual pruning, or finetuning, which is prohibitively expensive for large pretrained models.
\citet{kwon2022a} and \citet{ziplm} consider the more challenging task of post-training one-shot structured pruning, which is the main focus of our work. These consider pruning both heads in multi-head attention layers and intermediate dimension in FFN blocks.
\citet{kwon2022a} considers a local model based on the second-order (Hessian) information of the loss function to prune attention layer heads and hidden neurons in FFN.
Although effective, this approach can be prohibitively expensive in terms of runtime and/or memory for billion-parameter 
models as the ones we consider.
\citet{ziplm} consider a layer-wise reconstruction  error \citep{He2017,Luo17} to prune attention heads and hidden neurons in FFN. 
\citet{kwon2022a} and \citet{ziplm} consider BERT-Large and GPT2 models with sizes $<340$ million.

\noindent\textbf{Paper Organization.}
The rest of the paper is organized as follows. 
We present our problem formulation in Section \ref{sect:problem-formulation} and our proposed algorithm in Section \ref{sect:alg}. 
We provide empirical validation of our proposals on large vision and language models in Section \ref{sect:expt}.




\section{Problem Formulation}
\label{sect:problem-formulation}
In this section, we present \modelaname: a novel framework for structured pruning under latency constraints. Building on prior work in post-training structured pruning \cite{He2017,Luo17},  we adopt a layer-wise pruning strategy. This approach aims to selectively prune weights with minimal impact on performance at each layer, i.e., pruned weights \( w \) perform comparably to original pre-trained weights \( \widehat{w} \).

Formally, our approach minimizes the squared error loss between the output of a layer with \( \widehat{w} \) and the pruned weight \( w \) (to be learned) over \( N \) training samples \( \{X^{i}\}_{i=1}^N \). With $h(w,X)$ denoting the layer's output (before the activation function) with weight $w$ for input $X$, the loss reads:
\begin{equation}\label{eq:obj-ver1}
   L(w) =  \frac{1}{2N}\sum_{i=1}^N \left\|h(\widehat{w},X^{i})-h(w,X^{i})\right\|^2,
\end{equation}
subject to some structured sparsity constraints on $w$. 

In all models we consider, $h(w, X)$ is a linear function of $w$---the loss $L(w)$ is a quadratic function of \( w \) and can be expressed as
\begin{equation}\label{eq:obj-ver1.5}
    L(w) =  \frac{1}{2}w^\top H' w + G'^\top w.
\end{equation} 
Here, $H'$ and $G'$ are the quadratic and linear coefficients, respectively. $w$ is a vector with size that equals to the number of weights in the layer, potentially reaching hundreds of millions in LLMs (e.g.,  200 million weights in the OPT-30B model). This scale presents significant memory and computational challenges in saving $H'$ and evaluating $L(w)$.

To address the computational challenges, we make a key observation: the loss function \( L(w) \) is highly structured in the sense that $H'$ is highly sparse, and variables in $w$ can be divided into groups sharing the same quadratic coefficients. This insight allows us to represent $w$ as a matrix (denoted as $W$) and reformulate the loss as
\begin{equation}\label{eq:obj-ver2}
    L(W) =  \frac{1}{2}\text{Tr}(W^\top H W) + \text{Tr}(G^\top W).
\end{equation} 
This reformulation significantly downscales the problem, specifically the size of the quadratic coefficient matrix, from hundreds of millions to tens of thousands. For instance, with OPT-30B, we only need to handle a \( 27k \times 27k \) matrix $H$.



In the following discussion, we show how to represent the weight as a matrix $W$, the construction of matrices $H$ and $G$ in \eqref{eq:obj-ver2}, and the structured pruning constraints applied to $W$. This discussion covers dense layers, CNNs and LLMs. Furthermore,  we demonstrate that across all models, our problem can be uniformly formulated as a Mixed Integer Quadratic Programming (MIQP) problem.





\noindent\textbf{Structured pruning in dense layers.}
As a motivating example, we start with structured pruning in a dense (linear) layer with an input dimension $N_{in}$ and an output dimension $N_{out}$. In this case, we write $W$ as a $N_{in} \times N_{out}$ matrix, and the input $X$ over $N$ samples as a $N \times N_{in}$ matrix. The loss function can then be expressed as $L(W)=\frac{1}{2}\|X\widehat W - XW\|_F^2=\frac{1}{2}\Tr(W^\top (X^\top X) W) + \Tr( (X^\top X \widehat W)^\top W) + \frac{1}{2}\Tr(\widehat W^\top (X^\top X) \widehat W)$. Consequently, we set $H= X^\top X$ and $G=X^\top X \widehat W$ in \eqref{eq:obj-ver2}. For structured pruning within dense layers, our focus is on the removal of input neurons, which corresponds to pruning some certain rows in the matrix $W$.


\noindent\textbf{Structured pruning in CNNs.}
To prune the weights in a convolutional layer, we apply convolutional filters \( w \) of size \( C_{out} \times C_{in} \times k_H \times k_W \) to an input feature map \( X \) of dimensions \( C_{in} \times f_h \times f_w \). This feature map \( X \) is derived from a data point in the training set. The convolution of \( w \) with \( X \) produces an output matrix \( h(w,X) = \Conv\left(w, X\right) \) of size \( C_{out} \times f_h \times f_w \). In this notation, \( C_{in} \) and \( C_{out} \) are the number of input and output channels (the output channel also referred to as a filter), and \( k_h \), \( k_w \), \( f_h \), and \( f_w \) represent kernel and feature map dimensions, respectively. 



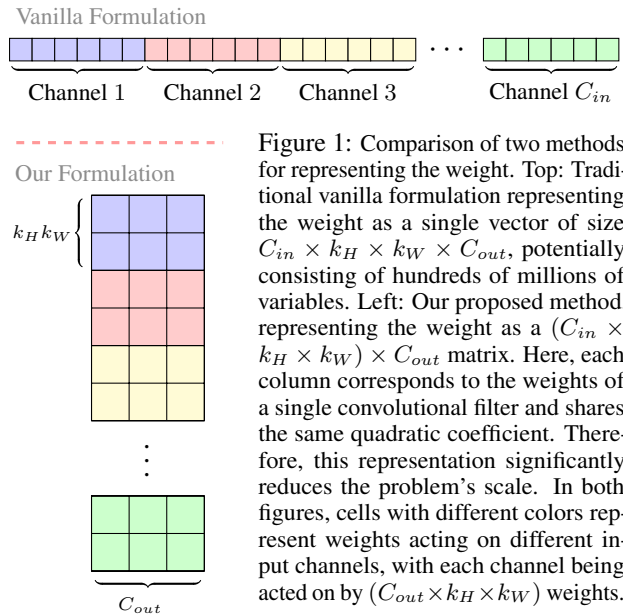
\begin{figure}[!b]
\centering

\begin{minipage}[c]{0.99\linewidth}
\centering
        \begin{tikzpicture}[scale=0.3]
  \node[text=gray, opacity=0.9] at (4.5,2) {\small {Vanilla Formulation}};
   \draw [draw=black,fill=blue!20] (0,0) rectangle (1,1);
  \draw [draw=black,fill=blue!20] (1,0) rectangle (2,1);
  \draw [draw=black,fill=blue!20] (2,0) rectangle (3,1);
  \draw [draw=black,fill=blue!20] (3,0) rectangle (4,1);
  \draw [draw=black,fill=blue!20] (4,0) rectangle (5,1);
  \draw [draw=black,fill=blue!20] (5,0) rectangle (6,1);
  \draw [draw=black,fill=red!20] (6,0) rectangle (7,1);
  \draw [draw=black,fill=red!20] (7,0) rectangle (8,1);
  \draw [draw=black,fill=red!20] (8,0) rectangle (9,1);
  \draw [draw=black,fill=red!20] (9,0) rectangle (10,1);
  \draw [draw=black,fill=red!20] (10,0) rectangle (11,1);
  \draw [draw=black,fill=red!20] (11,0) rectangle (12,1);
  \draw [draw=black,fill=yellow!20] (12,0) rectangle (13,1);
  \draw [draw=black,fill=yellow!20] (13,0) rectangle (14,1);
  \draw [draw=black,fill=yellow!20] (14,0) rectangle (15,1);
  \draw [draw=black,fill=yellow!20] (15,0) rectangle (16,1);
  \draw [draw=black,fill=yellow!20] (16,0) rectangle (17,1);
  \draw [draw=black,fill=yellow!20] (17,0) rectangle (18,1);
    \node at (19.5, 0.5) {\large{$\cdots$}};
  \draw [draw=black,fill=green!20] (21,0) rectangle (22,1);
  \draw [draw=black,fill=green!20] (22,0) rectangle (23,1);
  \draw [draw=black,fill=green!20] (23,0) rectangle (24,1);
  \draw [draw=black,fill=green!20] (24,0) rectangle (25,1);
  \draw [draw=black,fill=green!20] (25,0) rectangle (26,1);
  \draw [draw=black,fill=green!20] (26,0) rectangle (27,1);
  \draw [decorate, thick,
    decoration = {calligraphic brace}] (5.9,-0.25) --  (0.1,-0.25);
  \draw [decorate, thick,
    decoration = {calligraphic brace}] (11.9,-0.25) --  (6.1,-0.25);
  \draw [decorate, thick,
    decoration = {calligraphic brace}] (17.9,-0.25) --  (12.1,-0.25);
  \draw [decorate, thick,
    decoration = {calligraphic brace}] (26.9,-0.25) --  (21.1,-0.25);
  \node at (3,-1.4) {\small {Channel $1$}};
  \node at (9,-1.4) {\small {Channel $2$}};
  \node at (15,-1.4) {\small {Channel $3$}};
  \node at (24,-1.4) {\small {Channel $C_{in}$}};
  \end{tikzpicture}
\end{minipage}%
\\ \vspace{2mm}
    \begin{minipage}[c]{0.4\linewidth}
    
    \begin{tikzpicture}[scale=0.5]
   \draw[red!40, dashed,very thick] (-2,1.4) -- (3.5,1.4);
  \node[text=gray, opacity=0.9] at (0.1,0.6) {\small {Our Formulation}};
  \draw[step=1.0,draw=black,fill=blue!20] (0,0) grid (3.0,-2.0) rectangle (0,0);
         
  \draw[step=1.0,draw=black,fill=red!20] (0,-2.0) grid (3.0,-4.0) rectangle (0,-2.0);
  \draw[step=1.0,draw=black,fill=yellow!20] (0,-4.0) grid (3.0,-6.0) rectangle (0,-4.0);
  \node[rotate=90] at (1.5, -7) {\large{$\cdots$}};
  \draw[step=1.0,draw=black,fill=green!20] (0,-8.0) grid (3.0,-10.0) rectangle (0,-8.0);
  \draw [decorate, thick,
    decoration = {calligraphic brace}] (-0.25,-1.9) --  (-0.25,-0.1);
  \draw [decorate, thick,
    decoration = {calligraphic brace}] (2.9,-10.25) --  (0.1,-10.25);
   \node at (-1.3,-1) {\scriptsize {$k_Hk_W$}};
  \node at (1.3,-10.9) {\scriptsize {$C_{out}$}};
        \end{tikzpicture}
    \end{minipage}%
    \hfill
    \begin{minipage}[c]{0.59\linewidth}       
       \caption{\small{Comparison of two methods for representing the weight. Top: Traditional vanilla formulation representing the weight as a single vector of size \( C_{in} \times k_H \times k_W \times C_{out} \), potentially consisting of hundreds of millions of variables. 
       Left: Our proposed method, representing the weight as a \( (C_{in} \times k_H \times k_W) \times C_{out} \) matrix. Here, each column corresponds to the weights of a single convolutional filter and shares the same quadratic coefficient.  Therefore, this representation significantly reduces the problem's scale. In both figures, cells with different colors represent weights acting on different input channels, with each channel being acted on by $(C_{out} \times k_H \times k_W)$ weights.}}
       \label{fig:Wformulation}
    \end{minipage}
\end{figure}

An important insight here is that the outputs of any two filters are independent of each other's weight choices, and their weights share identical quadratic coefficients in \eqref{eq:obj-ver1}. Leveraging this point, we can represent the weight as a matrix \( W \) with dimensions \( (C_{in} \times k_H \times k_W) \times C_{out} \), where each column represents the weights of a single convolutional filter, as illustrated in Fig \ref{fig:Wformulation}.

This formulation allows us to express the loss $L(W)$ in the form \eqref{eq:obj-ver2}. Here, \( H \) is a positive semi-definite matrix with dimensions \( (C_{in}\times k_H \times k_W)\times(C_{in}\times k_H \times k_W) \), and \( G \) is the quadratic problem's linear term, with dimensions \( (C_{in}\times k_H \times k_W)\times C_{out} \). Both \( H \) and \( G \) can be computed using the training samples \( \{X^{i}\}_{i=1}^N \) and the original dense weight \( \widehat{W} \)  \footnote{The practical computation of \( H \) and \( G \) can be facilitated by, e.g., PyTorch's ``nn.unfold'' function}. 

\begin{figure}[!t]
    \centering
        \begin{tikzpicture}[scale=0.4]
   \node[align=center] at (3.5,8.3) {\scriptsize Feature map $X$};
   \node[align=center] at (3.4,7.5) {\scriptsize at layer $\ell$};
  \draw [draw=black,fill=green!20] (0,0) rectangle (0.5,5);
   \draw[draw=black,fill=green!20] (0.5,0) -- (2.5,2) -- (2.5,7) -- (0.5,5) -- cycle; 
   \draw[draw=black,fill=green!20] (0,5) -- (0.5,5) -- (2.5,7) -- (2,7) -- cycle;
   \fill (1.55, 3.5) circle (0.13cm);
   \fill (2.15, 3.5) circle (0.13cm);
   \fill (2.75, 3.5) circle (0.13cm);


   \fill[fill=yellow!20] (3,0) rectangle (3.5,5);
   \fill[fill=yellow!20] (3.5,0) -- (5.5,2) -- (5.5,7) -- (3.5,5) -- cycle; 
   \fill[fill=yellow!20] (3,5) -- (3.5,5) -- (5.5,7) -- (5,7) -- cycle;


   \draw[draw=black] (3,0) rectangle (3.5,5);
   \draw[draw=black] (3.5,0) -- (5.5,2) -- (5.5,7) -- (3.5,5) -- cycle; 
   \draw[draw=black] (3,5) -- (3.5,5) -- (5.5,7) -- (5,7) -- cycle;

   \fill[fill=gray!30,opacity=0.8] (3.5,0) rectangle (4,5);
   \fill[fill=gray!30,opacity=0.8] (4,0) -- (6,2) -- (6,7) -- (4,5) -- cycle; 
   \fill[fill=gray!30,opacity=0.8] (3.5,5) -- (4,5) -- (6,7) -- (5.5,7) -- cycle;

   \draw[draw=black,densely dotted] (3.5,0) -- (4,0); 
   \draw[draw=black,densely dotted] (3.5,5) -- (4,5); 
   \draw[draw=black,densely dotted] (5.5,7) -- (6,7); 
   
   \fill[fill=blue!20] (4,0) rectangle (4.5,5);
   \fill[fill=blue!20] (4.5,0) -- (6.5,2) -- (6.5,7) -- (4.5,5) -- cycle; 
   \fill[fill=blue!20] (4,5) -- (4.5,5) -- (6.5,7) -- (6,7) -- cycle;

   \draw[draw=black] (4,0) rectangle (4.5,5);
   \draw[draw=black] (4.5,0) -- (6.5,2) -- (6.5,7) -- (4.5,5) -- cycle; 
   \draw[draw=black] (4,5) -- (4.5,5) -- (6.5,7) -- (6,7) -- cycle;

  \node[align=center] at (9.75,6) {\scriptsize Weight $W$};
  \draw[step=0.5,draw=black,fill=blue!20] (9,5) grid (10.5,4) rectangle (9,5);
  \draw[step=0.5,draw=black,densely dotted,fill=gray!20] (9,4) grid (10.5,3) rectangle (9,4);
  \draw[step=0.5,draw=black,fill=yellow!20] (9,3) grid (10.5,2) rectangle (9,3);

   \fill (9.75, 1.7) circle (0.05cm);
   \fill (9.75, 1.5) circle (0.05cm);
   \fill (9.75,1.3) circle (0.05cm);
  \draw[step=0.5,draw=black,fill=green!20] (9,1) grid (10.5,0) rectangle (9,1);

   \draw [draw=black,fill=gray!50] (13,0) rectangle (15,5);
   \draw[draw=black,fill=gray!50] (15,0) -- (17,2) -- (17,7) -- (15,5) -- cycle; 
   \draw[draw=black,fill=gray!50] (13,5) -- (15,5) -- (17,7) -- (15,7) -- cycle;

   \node[align=center] at (15.6,8.3) {\scriptsize Feature map $Y$ };
   \node[align=center] at (15.4,7.5) {\scriptsize after convolution};
   
    \draw [draw=black!70,dashed, thick] (18,0) -- (18,7.5);
     \node [text=red!45] at (18,-0.5){\scriptsize ReLU};

    \draw [decorate, thick,
    decoration = {calligraphic brace}] (4.4,-0.25) --  (0.1,-0.25);
    \draw [decorate, thick,
    decoration = {calligraphic brace}] (14.9,-0.25) --  (13.1,-0.25);
    \node[align=center] at (2.25,-1) {\scriptsize $C_{in}$ };
    \node[align=center] at (14,-1) {\scriptsize $C_{out}$};
   
  \end{tikzpicture}
    \caption{
   Illustration of structured pruning in a convolutional layer to reduce feature map $X$'s width by pruning weights that act on certain input channels. This figure shows an example where weights acting on the second input channel (denoted by gray cells) are pruned.  Consequently, the corresponding channel (also in gray) in the feature map \( X \) becomes redundant and can be removed. 
    }
    \label{fig:CNN}
\end{figure}
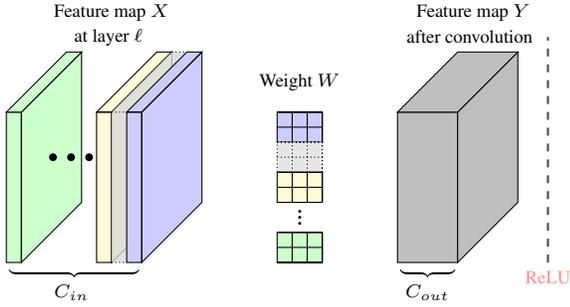

We adopt a channel pruning approach aimed at reducing the width of feature maps by selectively pruning weights that act on certain input channels, as depicted in Fig \ref{fig:CNN}. This method offers two benefits:~(i)~It enables the removal of redundant input channels affected by the pruned weights, along with the corresponding filters (from the previous layer) that generate these channels, thereby decreasing inference time. ~(ii)~Rather than removing entire convolutional filters, our approach prunes only parts of each filter. This selective pruning allows for the application of optimization techniques to the unpruned weights, aiming to closely replicate the output of the original filters by minimizing \( L(W) \).

As illustrated in Fig. \ref{fig:Wformulation}, each weight acting on a particular input channel corresponds to a set of rows in the weight matrix \( W \). Thus, to implement channel pruning, it suffices to prune all weights within some selected groups of rows in the matrix.

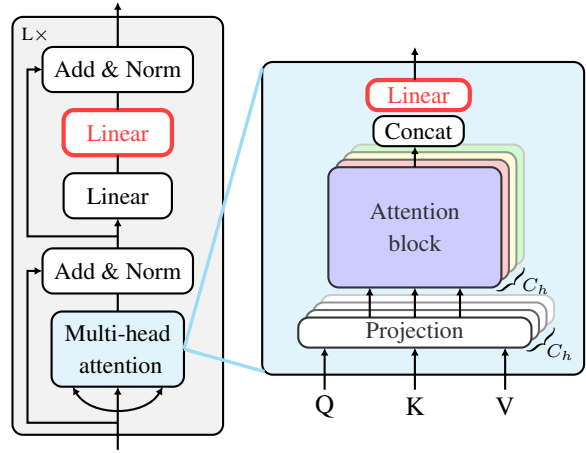
\begin{figure}[!t]
    \centering
        \begin{tikzpicture}[scale=0.4]

   \draw[rounded corners, thick,fill=gray!10] (0, -0.6) rectangle (7, 13.3);

   \draw[rounded corners, thick, fill=cyan!10] (1.3, 1) rectangle (5.7, 3.5) node[pos=.5,align=center] {\small{Multi-head }\\ \small{ attention}};

   \draw[rounded corners, thick,fill=white] (1, 4.1) rectangle (6, 5.6) node[pos=.5,align=center] {\small{Add \& Norm}};

   \draw[rounded corners, thick,fill=white] (1.7, 6.6) rectangle (5.3, 8.1) node[pos=.5,align=center] {\small{Linear}};

   \draw[rounded corners, ultra thick, red!75, fill=white] (1.7, 8.7) rectangle (5.3, 10.2) node[pos=.5,align=center] {\small{Linear}};

   \draw[rounded corners, thick,fill=white] (1, 10.8) rectangle (6, 12.3) node[pos=.5,align=center] {\small{Add \& Norm}};

   \draw[arrows = {-Latex[width=3pt, length=3pt]}, thick] (3.5,-1.1) -- (3.5,1);

   \draw[thick] (3.5,3.5) -- (3.5,4.1);
   \draw[arrows = {-Latex[width=3pt, length=3pt]}, thick] (3.5,5.6) -- (3.5,6.6);
   \draw[thick] (3.5,8.1) -- (3.5,8.7);
   \draw[thick] (3.5,10.2) -- (3.5,10.8);
   \draw[arrows = {-Latex[width=3pt, length=3pt]}, thick] (3.5,12.3) -- (3.5,13.8);
   \draw[arrows = {-Latex[width=3pt, length=3pt]}, thick] (3.5,-0.2) -- (0.5,-0.2) -- (0.5,4.85) -- (1,4.85);
   \draw[arrows = {-Latex[width=3pt, length=3pt]}, thick] (3.5,6) -- (0.5,6) -- (0.5,11.55) -- (1,11.55);

   \draw[arrows = {Latex[width=3pt, length=3pt]-Latex[width=3pt, length=3pt]}, thick] (2,1) to [bend right=60] (5,1);

   \node[align=center] at (0.85,12.7) {\scriptsize L$\times$ };

   \draw[rounded corners, thick,fill=cyan!10] (8.3, 1.4) rectangle (19, 11.8);

   \draw[rounded corners, thick, black!20, fill=white] (10.25, 3.05) rectangle (18, 4.05);
   \draw[rounded corners, thick, black!40, fill=white] (10., 2.8) rectangle (17.75, 3.8);
   \draw[rounded corners, thick, black!60, fill=white] (9.75, 2.55) rectangle (17.5, 3.55);
   \draw[rounded corners, thick, black!80, fill=white] (9.5, 2.3) rectangle (17.25, 3.3) node[pos=.5,align=center] {\small Projection};

   \draw[rounded corners, thick, black!20, fill=green!20] (11.25, 5.05) rectangle (17, 9.05);
   \draw[rounded corners, thick, black!40, fill=yellow!20] (11., 4.8) rectangle (16.75, 8.8);
   \draw[rounded corners, thick, black!60, fill=red!20] (10.75, 4.55) rectangle (16.5, 8.55);
   \draw[rounded corners, thick, black!80, fill=blue!20] (10.5, 4.3) rectangle (16.25, 8.3) node[pos=.5,align=center] {\small Attention\\ \small block};

    \draw[rounded corners, thick,fill=white] (12., 9) rectangle (15, 10.) node[pos=.5,align=center] {\small{Concat}};

    \draw[rounded corners, ultra thick, red!75,fill=white] (11.8, 10.2) rectangle (15.2, 11.2) node[pos=.5,align=center] {\small{Linear}};

    \draw[arrows = {-Latex[width=3pt, length=3pt]}, thick] (10.375,0.9) -- (10.375,2.3);
    \draw[arrows = {-Latex[width=3pt, length=3pt]}, thick] (13.375,0.9) -- (13.375,2.3);
    \draw[arrows = {-Latex[width=3pt, length=3pt]}, thick] (16.375,0.9) -- (16.375,2.3);
 
    \node[align=center] at (10.375,0.35) {Q};
    \node[align=center] at (13.375,0.35) {K};
    \node[align=center] at (16.375,0.35) {V};

    \draw[arrows = {-Latex[width=3pt, length=3pt]}, thick] (11.875,3.3) -- (11.875,4.3);
    \draw[arrows = {-Latex[width=3pt, length=3pt]}, thick] (13.375,3.3) -- (13.375,4.3);
    \draw[arrows = {-Latex[width=3pt, length=3pt]}, thick] (14.875,3.3) -- (14.875,4.3);

    \draw[arrows = {-Latex[width=3pt, length=3pt]}, thick] (13.375,8.3) -- (13.375,9);
    \draw[arrows = {-Latex[width=3pt, length=3pt]}, thick] (13.375,11.2) -- (13.375,12.3);

    \draw[cyan!35,line width=1.3pt] (5.7,2.25) -- (8.3,11.6);
    \draw[cyan!35,line width=1.3pt] (5.7,2.25) -- (8.38,1.5);

    \draw [decorate, thick,
     decoration = {calligraphic brace}] (18,2.95) -- (17.1,2.2) ;
    \draw [decorate, thick,
     decoration = {calligraphic brace}] (17,4.95) -- (16.1,4.2) ;

    \node[align=center] at (18.2,2.15) {\scriptsize $C_{h}$};
\node[align=center] at (17.4,4.35) {\scriptsize $C_{h}$};
     
  \end{tikzpicture}
    \caption{Illustration of structured pruning to accelerate a Transformer layer. The multi-head attention comprises $C_h$ distinct single-head attention blocks. The outputs from these blocks are concatenated and processed through a linear sublayer.
    We prune the linear sublayer for multi-head integration and the second sublayer of the feed-forward network, as marked in red. }
    \label{fig:LLM}
\end{figure}

\noindent\textbf{Structured pruning in LLMs.}
We now discuss structured pruning in Transformers \citep{zhang2022opt}. Figure \ref{fig:LLM} presents the structure of a layer in a Transformer. In line with previous studies \cite{kwon2022a,ziplm}, our focus is on two structural changes: removing attention heads and reducing the intermediate dimension in fully connected layers. This involves pruning the last linear sublayer in multi-head attention and the second sublayer of the feed-forward network, as highlighted in red in Fig \ref{fig:LLM}. 

Given that these two sublayers are linear, we can represent \( W \) as a $N_{in}\times N_{out}$ matrix, where $N_{in}$ and $N_{out}$ are the input and output dimensions of the linear layer, respectively. Following the discussion on structured pruning in dense layers, this allows us to express the loss \( L(W) \) in the format shown in \eqref{eq:obj-ver2}.

We now detail the structured pruning constraints for these two types of structural removal:
\begin{enumerate}[noitemsep,topsep=0pt,parsep=0pt,partopsep=0pt, leftmargin=*]
    \item  Removing attention heads: the output of each attention head is a \( D_{head} \)-dimensional vector, feeding into the multi-head integration linear sublayer. To prune an attention head, we remove \( D_{head} \) consecutive rows from the linear sublayer's weight matrix \( W \). 
    \item  Reducing the intermediate dimension in fully connected layers: this involves removing neurons from hidden states, equivalent to removing individual rows from the second linear sublayer's weight matrix. This case aligns with structured pruning for dense layers.
\end{enumerate}

\noindent\textbf{Structured pruning as a MIQP.}
We now introduce a unifying framework incorporating all the models discussed previously, demonstrating that structured pruning for these models can be formulated as a mixed-integer quadratic programming (MIQP) problem.

We first divide the rows of the weight matrix \( W \) into partitions \( Q^1, Q^2, \dots, Q^{p} \). To determine $p$ and the partition, we consider three cases:
\begin{enumerate}[noitemsep,topsep=0pt,parsep=0pt,partopsep=0pt, leftmargin=*]
    \item  Dense layers:  \( p = N_{in} \) (input dimension) and each \( Q^j = \{j\} \) denotes weights for the $j$-th input neuron.
    \item Convolutional layers: $p=C_{in}$ (the number of input channels) and each partition \( Q^j \) includes rows \( \{(j-1)k_Hk_W + 1, \dots, jk_Hk_W\} \), represents weights acting on the \( j \)-th input channel.
    \item  Attention heads: \( p = C_h \) (the number of attention heads) and each partition \( Q^j = \{(j-1)D_{head} + 1, \dots, jD_{head}\} \), represents weights for the $j$-th head.    
\end{enumerate}

We use a binary variable $z_j$ to indicate whether weights in $Q_j$ are kept or pruned. \modelaname~aims to prune \( p' \) groups of weights while minimizing the loss $L(W)$, and can be described by the following problem (here $\wbf_i$ denotes the $i$-th row of matrix $W$) with both discrete ($z$) and continuous ($W$) variables:
\begin{equation}\label{eq:MIQP}
\begin{aligned}
   \min_{W,z}\,\,\,\, &  L(W):=\frac{1}{2}\text{Tr}(W^\top H W) - \text{Tr}(G^\top W) \\
    \text{s.t. }\,\,   &    \wbf_i \cdot (1-z_j)=\textbf{0},\,\,\forall\,j\in[p],\,i\in Q^j\\
    & \,\sum_{j=1}^p z_j = p-p',\,\,\,z_j\in\{0,1\}.
\end{aligned}
\end{equation}



\section{Algorithm}\label{sect:alg}

The main goal of this paper is one-shot structured pruning on large-scale models. It is important to obtain high-quality solutions to the MIQP derived from structured pruning both effectively, due to the large scale of the problem, and accurately since we only prune once and will not fine-tune the weights via SGD. To this end, we first reformulate the MIQP \eqref{eq:MIQP} (involving both discrete and continuous variables) to a combinatorial problem in 
Section~\ref{sect:convert} (with only discrete variables). We then introduce an efficient approximate solver based on local search for achieving a high-quality solution in Section \ref{sect:algcomb}.

\subsection{A combinatorial optimization reformulation}\label{sect:convert}

The MIQP problem \eqref{eq:MIQP} involves two sets of variables: the binary variables \( z \) and the weight matrix \( W \). In practice, $z$ includes up to a few thousand variables, while the weight matrix $W$ may have millions of variables.
Fixing the binary variables \( z \) reduces the problem to a quadratic one, which, despite its large scale, is relatively simple to solve optimally. However, the critical task is to wisely determine the values of \( z \), as they indicate which portions of a layer's input are retained and how effectively the output of the dense weight can be approximated by the pruned weight. 

In light of this, we reformulate the MIQP problem as a combinatorial problem, primarily focusing on optimizing \( z \). We denote the set of indices where \( z \) is zero as \( S \), and define \( f(S) \) as the minimum loss \( L(W) \) achieved by pruning all rows in \( Q^j \) for \( j \) in \( S \) from the weight matrix \( W \) and then updating the remaining weights:
\begin{equation}\label{eq:deff}
\begin{aligned}
   f(S)& =\min_{W}\,\, L(W):=\frac{1}{2}\text{Tr}(W^\top H W) - \text{Tr}(G^\top W) \\
       &\quad\,\,\,\,\, \text{s.t. } \,\,\wbf_i=0,\,\, \forall\,i\in Q^j,\,j\in S.
\end{aligned}
\end{equation}
The objective is to identify a set \( S \) with \( p' \) elements that minimizes \( f(S) \), and can be written as the following combinatorial problem:
\begin{equation}\label{eq:struc}
    \min_{S\subset \{1, 2, \dots, p\}} \quad f(S), 
    \qquad\text{s.t. } \quad |S|=p'. 
\end{equation}
Optimizing \( f(S) \) remains a challenging problem, as potential \( S \) selections grow exponentially with the number of channels. For example, pruning half of 32 channels in a CNN layer results in over 600 million possible choices for \( S \). Moreover, calculating the value of \( f(S) \) for a specific \( S \) involves solving a quadratic problem with up to millions of variables, further complicating this problem. In the next section, we introduce a highly efficient method for identifying a high-quality \( S \).

\subsection{Local search based approximate solver}\label{sect:algcomb}

To motivate our local search based approximate solver, we first discuss calculating \( f(S) \) for a specific set \( S \). Define
\begin{equation}\label{eq:indexS}
    I_S:=\left\{i\mid i\in Q^j \text{ for some } j \notin S\right\}
\end{equation}
as the set of rows not constrained to zero in \eqref{eq:deff}. Consequently, computing the value of \( f(S) \) becomes a quadratic problem focusing on the weights in rows from \( I_S \):
\begin{equation}\label{eq:deff2}
   f(S) =\min_{W_{I_S}}\,\, \frac{1}{2}\text{Tr}(W_{I_S}^\top H_{I_S,I_S} W_{I_S}) - \text{Tr}(G_{I_S}^\top W_{I_S}) 
\end{equation}
Here, \( H_{I_S,I_S} \) is the submatrix of \( H \) with rows and columns in \( I_S \), and \( W_{I_S} \,(G_{I_S}) \) is the submatrix of \( W\,(G) \) with rows in \( I_S \). The optimal value of this quadratic problem is
\begin{equation}\label{eq:optf}
    f(S) = -\frac{1}{2}\Tr\left(G_{I_S}^\top (H_{I_S,I_S})^{-1}G_{I_S}\right).
\end{equation}
The time complexity of computing \eqref{eq:optf} is \( O(d_1^2(d_1+d_2)) \), assuming \( W \) is a \( d_1 \times d_2 \) matrix. Solving the combinatorial optimization problem \eqref{eq:struc} requires evaluating \( f(S) \) for numerous distinct sets \( S \). In LLMs, the scales of \( d_1 \) and \( d_2 \) can reach up to tens of thousands, presenting substantial computational challenges.

To accelerate the evaluation of $f(S)$, a crucial insight is that if we already have \( f(S') \) computed for a set \( S' \) similar to \( S \), we can apply low-rank matrix updates to efficiently calculate \( f(S) \). This leads to the following result:
\begin{proposition}\label{thm:complexity}
    Given two sets \( S \) and \( S' \). Suppose we have computed the value of \( f(S') \), the inverse of \( H_{I_{S'},I_{S'}} \) and the optimal weight matrix for \( f(S') \). The value of \( f(S) \), the inverse of \( H_{I_{S},I_{S}} \) and the optimal weight matrix for $f(S)$ can then be computed within \( O(td_1(d_1+d_2)) \) time complexity, where \( t = |I_S\Delta I_{S'}| \) denotes the symmetric difference between \( I_S \) and \( I_{S'} \).
\end{proposition}
The proof of Proposition \ref{thm:complexity} is in Section \ref{sect:prooftc}. It inspires us to  perform a local search at each iteration around the current solution $S'$ by approximately solving:
\begin{equation}\label{eq:local}
    \min_{S}\,\, f(S)\quad
    \text{s.t.}\,\, |S\Delta {S'}|\le \hat t,\,\,|S|\ge |S'|+\hat p. 
\end{equation}
We restrict the symmetric difference between $S'$ and $S$ by $\hat t$. This constraint allows efficient computation of compute \( f(S) \), the inverse of \( H_{I_{S},I_{S}} \) and the optimal weight matrix for \( f(S) \), as per Proposition \ref{thm:complexity}. We require $|S|\ge |S'|+\hat p$ since our strategy is to start with the empty set and gradually increase the cardinality of $S$, which means we begin with the dense weight and gradually prune the structural components in the neural network.

To solve \eqref{eq:local}, we replace elements in $S'$ with minimal impact on the objective with those that have more significant effects.
The importance of each element in \([p]\) is evaluated based on the objective's change when the element is added to or removed from \( S' \).
 This is formally expressed as:
\begin{equation}\label{eq:impact}
     B_{j}=\left\{ \begin{array}{ll}
      f\left(S'\right) - f\left(S'\setminus\{j\}\right),    & \text{ if }j \in S' \vspace{1.2mm} \\
      f\left(S'\cup\{j\}\right) - f\left(S'\right),    & \text{ otherwise, } 
     \end{array} \right.\,\,\forall j\in [p].
\end{equation}
We set \( s_1 = \lfloor \frac{\hat t-\hat p}{2}\rfloor \) and \( s_2 = \lfloor \frac{\hat t+\hat p}{2}\rfloor \). \( S_{out} \) includes the \( s_1 \) elements in \( S' \) with the lowest \( B_j \) values, while \( S_{in} \) includes the \( s_2 \) elements in \( [p]-S' \) with the highest \( B_j \) values. Then, we derive the solution to \eqref{eq:local} as:
\begin{equation}
    S = (S' \setminus S_{out}) \cup S_{in}.
\end{equation}

\begin{algorithm}[!t]
\setstretch{1.1}
\begin{algorithmic}[1]
\REQUIRE The number of iterations $T$, two lists $\{t_i\}_{i=1}^T$ and $\{p_i\}_{i=1}^T$ with $\sum_{i=1}^Tp_i=p'$ and $p_i\le t_i,\forall i\in[T]$.
\STATE Initialize with $S_0=\emptyset$, compute $H^{-1}$ and the optimal weight $W^*=H^{-1}G$.
\FOR{$i=1,2,\dots,T$}
\STATE Conduct a local search on $S_{i-1}$: solve problem \eqref{eq:local} with $S'=S_{i-1}$, $\hat t=t_i$ and $\hat p=p_i$ and get $S_i$.
\STATE Compute $f(S_i)$, the inverse of \( H_{I_{S_i},I_{S_i}} \) and the optimal weight matrix for \( f(S_i) \).
\ENDFOR
\STATE \textbf{Output:} The set $S_T$ that approximately optimizes \eqref{eq:struc}.
\end{algorithmic}
\caption{Local search-based approach for solving \eqref{eq:struc}.}
\label{alg:localsearch}
\end{algorithm}

The method we propose is concisely outlined in Algorithm~\ref{alg:localsearch}. Proposition \ref{thm:alg} details the time and memory complexities of 
Algorithm \ref{alg:localsearch}, with its proof provided in Section \ref{sect:proofalg}. 

\begin{proposition}\label{thm:alg}
    Assume each group in the row partition $Q^1, Q^2\dots, Q^k$ is of equal size. Then, Algorithm \ref{alg:localsearch} can be executed in $O(Td_1^2(d_1+d_2))$ time and requires $O(d_1(d_1+d_2))$ memory. Furthermore, if we choose $t_i=p_i$ for all $i\in [T]$, then the time complexity reduces to $O(d_1^2(d_1+d_2))$.
\end{proposition}

\begin{figure}[!b]
    \centering
        \begin{tikzpicture}[scale=0.44]

  \draw[thick, fill=green!10] (1,0) ellipse (4cm and 2cm);
  \draw[thick, fill=red!10] (0.5,0) ellipse (3cm and 1.5cm);
  \draw[thick, fill=blue!10] (0,0) ellipse (2cm and 1cm);

   \node[align=center] at (1.2,0) {\scriptsize $S_1$};
   \node[align=center] at (2.7,0) {\scriptsize $S_2$};
   \node[align=center] at (4.2,0) {\scriptsize $S_3$};
   
  \draw[thick, fill=blue!10] (10,0) ellipse (3cm and 1.5cm);
  \draw[thick] (11.5,0) ellipse (3cm and 1.5cm);
  \draw[thick, fill=red!50,opacity=0.2] (11.5,0) ellipse (3cm and 1.5cm);

  \node[align=center] at (7.9,0) {\scriptsize $S_1$};
  \node[align=center] at (13.6,0) {\scriptsize $S_2$};

  \end{tikzpicture}
    \caption{The structure of sets $\{S_i\}_{i=1}^T$ under different choices of $\hat p$. Left: with a large $\hat p$, Algorithm \ref{alg:localsearch} mimics a greedy pruning approach, incrementally expanding the set $S$ and resulting in nested sets $S_1\subset S_2\subset S_3\cdots$. Right: with a small $\hat p$, Algorithm \ref{alg:localsearch} employs a local swapping strategy, leading to sets without a nested structure.}
\label{fig:step}
\end{figure}
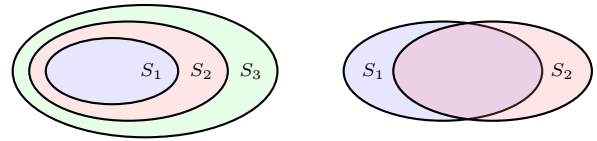

We now discuss the effects of parameters \(\hat t\) and \(\hat p\) in \eqref{eq:local} on the performance of Algorithm \ref{alg:localsearch}. The parameter \(\hat t\) determines the range of the local search. A larger \(\hat t\) enables more aggressive updates yet complicates solving \eqref{eq:local}, reducing the accuracy of our approximate solution to it. Conversely, a smaller \(\hat t\) leads to precise updates, but this conservative approach requires much more iterations for convergence. As for \(\hat p\), its larger values (e.g., $\hat t=\hat p$) make Algorithm \ref{alg:localsearch} resemble a greedy pruning method. In this mode, each iteration adds elements to \( S \) with the greatest impact on the objective \( f(S) \). On the other hand, a smaller \(\hat p\) shifts the algorithm towards a local swapping strategy, where elements within \( S \) are exchanged with more impactful ones from outside \( S \) to find a set with a smaller objective value. Figure \ref{fig:step} provides an illustration of Algorithm \ref{alg:localsearch} under different $\hat p$ settings.

In our experiments, we found that the performance of our proposed algorithm is not sensitive to the choice of $\hat t$ and $\hat p$. Therefore, we do not tune over these parameters much, and we set \(\hat t = \hat p \leq 10\), with the precise values of \(\hat t\) and \(\hat p\) depending on the problem size (detailed in Section \ref{sect:expt-setup}). We also conduct an ablation study to examine the algorithm’s performance with various choices of $\hat t$ and $\hat p$ and show the low sensitivity of our algorithm to these hyperparameters. The results are presented in Appendix \ref{sect:choicetk}.

\section{Numerical Experiments}\label{sect:expt}
In this section, we compare our proposed framework \modelaname~with leading structured pruning methods in vision and language models. We provide detailed information on the experimental setup and reproducibility in Appendix~\ref{sect:expt-setup}. Additional experimental results and ablation studies are given in Appendix \ref{sect:addexp}.

Our proposed framework \modelaname~employs a one-shot pruning method, which prunes weights just once without finetuning. 
On the other hand, some structured pruning methods \cite{Li2017,Sui2021,Luo17} perform a fine-tuning procedure after pruning to retrain the weights via stochastic gradient descent. 
One-shot pruning is more efficient than fine-tuning, which typically requires a significantly larger number of training samples and greater computational resources. For instance, fine-tuning ResNet50 on ImageNet involves over a million training samples and days of training, whereas our one-shot pruning experiments only take a few minutes and 500 training samples.
Additionally, fine-tuning large language models often becomes impractical due to the high demand for computational resources. Therefore, in our experiments, we focus on one-shot pruning.
To ensure fairness in our comparisons, we also consider the performance of competing methods in one-shot setting. 

\subsection{Structured pruning in vision models}
We consider various pre-trained convolutional networks including ResNet20~(\citealp{he2016deep}, 260k parameters) trained on CIFAR10~\citep{krizhevsky2009learning}, MobileNet (\citep{howard2017mobilenets}, 4.2M parameters) and ResNet50 (25.6M parameters) trained on ImageNet~\citep{deng2009imagenet}. We assess the performance of all structured pruning methods using $500$ training samples as the calibration dataset.

\noindent\textbf{Competing methods.} We compare it with several one-shot pruning methods in vision models: (i) Magnitude Pruning (MP)~\citep{mozer1989using,He2018a}, (ii) CHIP~\cite{Sui2021}, (iii) FPGM~\cite{He2018geometric}, (iv) Lasso~\cite{He2017} and (v) ThiNet~\cite{Luo17}. The configuration details for these methods and the parameter settings for \modelaname~are outlined in Section \ref{sect:cnnsetup}.



\begin{table}[!t]
    \centering
\caption{One-shot structured pruning performance (accuracy) of various methods on ResNet20, MobileNetV1, and ResNet50. The speedup ratio denotes the inference time improvement of pruned models over dense models. For all methods, we take ten runs and report the mean.}

\label{tab:accuracy}
    \resizebox{0.99\columnwidth}{!}
    {\begin{tabular}{c|c|ccccc|c}
    \toprule
\footnotesize Model & Speedup & MP & CHIP & FPGM & Lasso & ThiNet & \modelaname \\
\midrule
\multirow{6}{*}{
\begin{minipage}{2cm}
\begin{center}
    ResNet20\\
    on CIFAR10\\
    (91.36\%)
\end{center}
\end{minipage}
}&1.3x & 55.37 & 67.24 & 39.01 & 89.03 & 89.16 & \textbf{89.20} (±0.12)\\ 
&1.4x & 39.54 & 39.05 & 24.37 & 86.68 & 87.09 & \textbf{87.70} (±0.16)\\ 
&1.7x & 26.92 & 19.16 & 12.82 & 83.55 & 84.11 & \textbf{85.05} (±0.19)\\ 
&2.0x & 18.98 & 11.81 & 10.92 & 78.80 & 79.30 & \textbf{81.10} (±0.49)\\ 
&2.6x & 11.04 & 9.97 & 12.45 & 68.58 & 70.35 & \textbf{72.60} (±0.78)\\ 
&3.5x & 10.21 & 10.31 & 12.60 & 57.15 & 58.68 & \textbf{61.82} (±0.98)\\
\midrule
\multirow{6}{*}{
\begin{minipage}{2cm}
\begin{center}
    MobileNetV1\\
    on ImageNet\\
    (71.95\%)
\end{center}
\end{minipage}
}&1.3x & 21.66 & 33.16 & 4.42 & 70.13 & 70.07 & \textbf{70.70} (±0.05)\\ 
&1.4x & 1.60 & 5.36 & 0.58 & 67.65 & 67.66 & \textbf{68.89} (±0.08)\\ 
&1.5x & 0.13 & 0.77 & 0.11 & 63.79 & 64.06 & \textbf{66.37} (±0.15)\\ 
&1.7x & 0.10 & 0.22 & 0.10 & 58.88 & 59.51 & \textbf{63.03} (±0.31)\\ 
&1.9x & 0.10 & 0.10 & 0.13 & 52.72 & 53.34 & \textbf{58.39} (±0.36)\\ 
&2.2x & 0.10 & 0.10 & 0.10 & 45.47 & 45.77 & \textbf{52.07} (±0.57)\\  
\midrule
\multirow{6}{*}{
\begin{minipage}{2cm}
\begin{center}
    ResNet50\\
    on ImageNet\\
    (77.01\%)
\end{center}
\end{minipage}
}&1.2x & 60.57 & 57.06 & 59.53 & 73.82 & 73.49 & \textbf{74.33} (±0.08)\\ 
&1.3x & 9.28 & 16.86 & 18.79 & 67.07 & 66.29 & \textbf{69.30} (±0.16)\\ 
&1.4x & 3.28 & 5.34 & 5.48 & 61.58 & 60.44 & \textbf{65.32} (±0.27)\\ 
&1.6x & 0.95 & 1.24 & 1.42 & 53.15 & 51.91 & \textbf{59.14} (±0.44)\\ 
&1.7x & 0.37 & 0.57 & 0.40 & 41.68 & 41.08 & \textbf{50.16} (±0.59)\\ 
&1.9x & 0.25 & 0.29 & 0.21 & 29.23 & 28.87 & \textbf{38.45} (±0.63)\\ 
\bottomrule
 \end{tabular} }
\end{table}

\begin{figure*}[!h]
\small
\centering
\begin{tabular}{ccc}
    \includegraphics[width=0.3\textwidth]{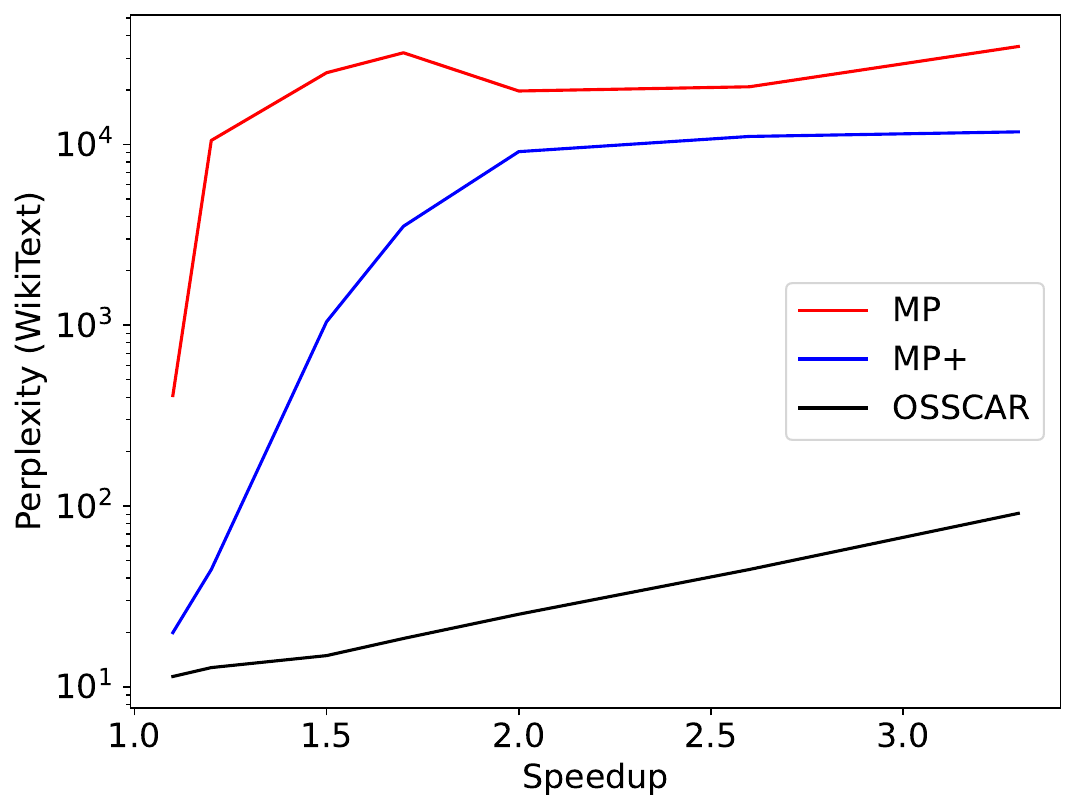}&  \includegraphics[width=0.3\textwidth]{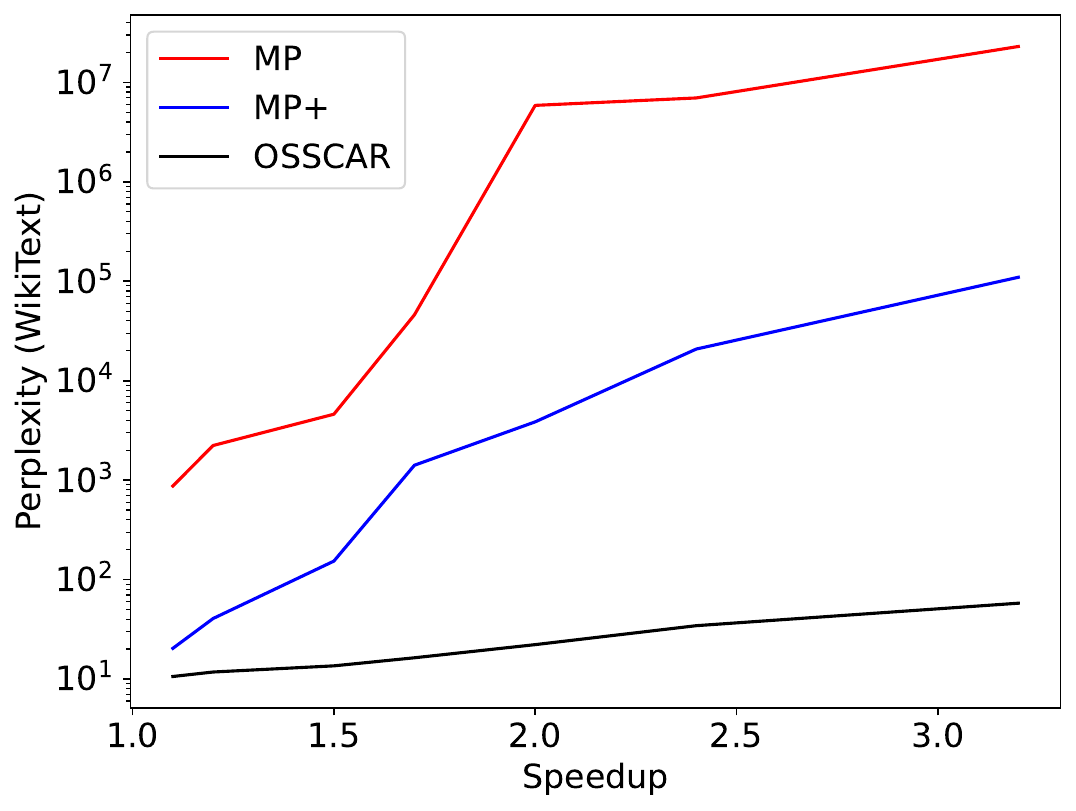}&  \includegraphics[width=0.3\textwidth]{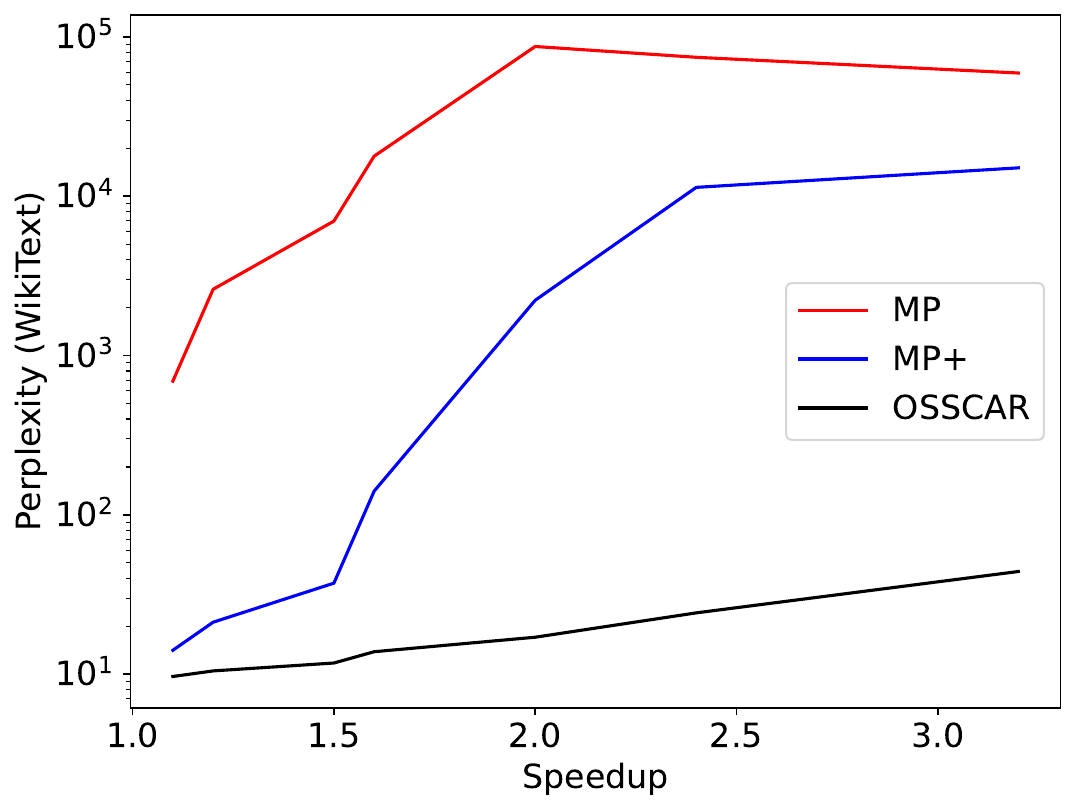}\\
    OPT-6.7B & OPT-13B & OPT-30B
    \end{tabular}
\caption{Perplexity performance on WikiText (in log-scale) for one-shot structured pruning of OPT models (6.7B, 13B, and 30B). The speedup ratio denotes the inference time improvement of pruned models over dense models. For all methods, we take ten runs and report the mean perplexity.
}  
\label{fig:llms-opt-large}
\end{figure*}

\noindent\textbf{Accuracy performance.} 
Table \ref{tab:accuracy} compares the test accuracy for pruned ResNet20, MobileNetV1, and ResNet50 across different speedup ratios. \modelaname~consistently achieves higher accuracy compared to existing methods, particularly at higher speedup ratios. Also, \modelaname~achieves $\sim20$\% lower loss in solving a single-layer pruning problem compared to existing methods (see Section \ref{sect:addexp}).

\begin{table}[!b]
\caption{
Perplexity performance on Wikitext for one-shot structured pruning of OPT models (1.3B, 2.7B, and 6.7B). The speedup ratio denotes the inference time improvement of pruned models over dense models. For all methods we take ten runs and report the mean and standard error.
}
\label{tab:osscar-llm}
\resizebox{\columnwidth}{!}{%
\begin{tabular}{c|c|ccc|c}
\toprule
 Model & Speedup & MP & MP+ & ZipLM & \modelaname \\ \midrule
\multirow{7}{*}{OPT-1.3B } & 1.2x & 163.0 ($\pm$0.20) & 22.01 ($\pm$0.09) & \textbf{14.58} ($\pm$0.08) & 15.54 ($\pm$0.15) \\
 & 1.3x & 1834 ($\pm$1.00) & 41.46 ($\pm$0.28) & 61.36 ($\pm$5.19) & \textbf{17.53} ($\pm$0.11) \\
 & 1.4x & 7412 ($\pm$29.0) & 4953 ($\pm$151) & 729.0 ($\pm$114) & \textbf{20.49} ($\pm$0.36) \\
 & 1.7x & 8752 ($\pm$55.0) & 4802 ($\pm$287) & 1829 ($\pm$66.0) & \textbf{25.74} ($\pm$0.54) \\
 & 2.0x & 8439 ($\pm$15.0) & 6490 ($\pm$113) & 3529 ($\pm$346) & \textbf{37.87} ($\pm$0.64) \\
 & 2.6x & 9546 ($\pm$4.00) & 7608 ($\pm$241) & 6424 ($\pm$301) & \textbf{68.50} ($\pm$1.89) \\
 & 3.3x & 12006 ($\pm$50.0) & 8772 ($\pm$256) & 9424 ($\pm$761) & \textbf{153.0} ($\pm$3.56) \\ \midrule
\multirow{7}{*}{OPT-2.7B} & 1.2x & 234.2 ($\pm$3.07) & 22.33 ($\pm$0.12) & \textbf{12.14} ($\pm$0.03) & 13.14 ($\pm$0.17) \\
 & 1.3x & 3817 ($\pm$108) & 46.31 ($\pm$0.59) & 21.83 ($\pm$3.47) & \textbf{14.94} ($\pm$0.22) \\
 & 1.4x & 6957 ($\pm$794) & 111.2 ($\pm$2.50) & 414.9 ($\pm$113) & \textbf{17.11} ($\pm$0.21) \\
 & 1.7x & 13903 ($\pm$208) & 9977 ($\pm$5025) & 1820 ($\pm$422) & \textbf{21.18} ($\pm$0.24) \\
 & 2.0x & 12793 ($\pm$48.0) & 12003 ($\pm$2419) & 3611 ($\pm$840) & \textbf{29.48} ($\pm$0.46) \\
 & 2.4x & 11975 ($\pm$18.0) & 15979 ($\pm$3180) & 9209 ($\pm$2356) & \textbf{51.90} ($\pm$1.06) \\
 & 3.0x & 15874 ($\pm$62.0) & 12433 ($\pm$596) & 14039 ($\pm$1863) & \textbf{102.7} ($\pm$3.24) \\ 
 \bottomrule
\end{tabular}%
}
\end{table}

\subsection{Structured Pruning in Large Language Models}
Next, we evaluate the usefulness of our proposed framework \modelaname~for one-shot structured pruning on LLMs.

\noindent\textbf{Models and datasets.}
We focus on pruning the OPT model family \citep{zhang2022opt}, with sizes ranging from 1.3 billion to 30 billion parameters.  
For calibration data, we adopt the approach of \citealp{pmlr-v202-frantar23a}, utilizing 128 segments of 2048 tokens each, randomly selected from the first shard of the C4 dataset \citep{Raffel2019}. We focus on perplexity as our metric, recognized for its challenge and stability in evaluating performance of pruned models \citep{Yao2022,Xiao2023,pmlr-v202-frantar23a}. The perplexity is calculated following precisely the procedure described by HuggingFace \citep{Perplexity}, using full stride.
We consider the test sets of raw-WikiText2 \citep{merity2017pointer} and PTB \citep{Marcus1994} as well as a subset of the C4 validation data, all popular benchmarks in LLM pruning literature \citep{Yao2022,Xiao2023,pmlr-v202-frantar23a}. We use the HuggingFace Transformers library \citep{wolf-etal-2020-transformers} for handling the models and datasets.




\noindent\textbf{Competing methods.} We compare against multiple one-shot structured pruning methods: (i) Magnitude pruning (MP), (ii) Magnitude pruning with a layer-wise refinement step (MP+), and (iii) ZipLM\footnote{The authors of ZipLM (state-of-the-art) only considered models of sizes $<340$ million. We apply the ZipLM framework to OPT models. For the second stage of their framework (structured SPDY search), we turn off the randomized search component, as it appears to be computationally intractable.} \citep{ziplm}.  The configuration details for these methods, along with the parameter settings for \modelaname, are outlined in Section \ref{sect:llmsetup}.

\noindent\textbf{Results.} We report the perplexity performance on the raw-WikiText2 test set across various speedup ratios for different models (OPT-1.3B, OPT-2.7B), as shown in Table \ref{tab:osscar-llm}. 
Interestingly, we observe a rapid increase in perplexity for pruned models using baseline methods, even at relatively small speedup ratios.  
\modelaname, in contrast, demonstrates substantial gains over baseline methods. Notably, it shows a much smaller increase in perplexity with higher speedups. For instance, \modelaname~achieves a perplexity 125 times lower than the state-of-the-art ZipLM framework for one-shot structured pruning for the OPT 2.7 billion parameter model at a 2x inference speedup. 



\paragraph{Performance on larger OPT Models}
Next, we showcase the one-shot pruning performance of \modelaname~on larger OPT Models, specifically the 6.7B, 13B, and 30B versions. ZipLM \citep{ziplm} is unable to process these models due to memory limitations\footnote{The original ZipLM pipeline runs out of memory for 1.3B and 2.7B models as well. We added CPU offloading (see Section \ref{sect:llmsetup}), allowing it to work with 1.3B and 2.7B models.}. Therefore, our framework is compared solely against magnitude pruning-based methods (MP and MP+). We display the perplexity results for these models on raw-WikiText2 in Figure \ref{fig:llms-opt-large}. The logarithmic scale of the y-axis clearly demonstrates that \modelaname~can reduce the perplexity by at least two orders of magnitude compared to the existing baselines in such large models.

The overall loss in perplexity from one-shot structured pruning can be mitigated by increasing the sample size from 128 to 2048. For instance, at a 2x speedup, this adjustment can reduce perplexity by up to 3.5 points. We explored this in an ablation study detailed in the Supplementary Section \ref{sect:llmsize}.


\noindent\textbf{Timing comparison.}
Our framework stands out for its time efficiency in pruning networks. Table \ref{tab:llm-timing-comparison} displays the total time different algorithms take to prune a model to various speed-up ratios. As shown, \modelaname~is notably faster, achieving speeds $6$--$8$ times faster than the ZipLM framework. Remarkably, our optimization framework is only 1.5x slower than MP+ baseline, yet it offers substantial improvements in predictive performance.



\begin{table}[!h]
\caption{Total time taken by different algorithms to prune a model to various speed-up ratios. For all methods we take ten runs and report the mean and standard error.}
\label{tab:llm-timing-comparison}
\resizebox{\columnwidth}{!}{%
\begin{tabular}{ccccc}
\toprule
Model & MP & MP+ & ZipLM & \modelaname \\ \midrule
\multicolumn{1}{c|}{OPT-1.3B} & 0.469 ($\pm$0.00) & 1166 ($\pm$3.00) & 9161 ($\pm$670) & 1455 ($\pm$4.00) \\ 
\multicolumn{1}{c|}{OPT-2.7B} & 0.996 ($\pm$0.00) & 2229 ($\pm$8.00) & 22839 ($\pm$999) & 2842 ($\pm$12.0) \\ 
\multicolumn{1}{c|}{OPT-6.7B} & 2.551 ($\pm$0.01) & 5084 ($\pm$67.0) & - & 7434 ($\pm$96.0) \\  \bottomrule
\end{tabular}%
}
\end{table}



\section{Conclusion}
We introduce \modelaname, a novel optimization framework for one-shot structured pruning in large-scale vision and language models.
\modelaname is based on a reformulation of the layer-wise reconstruction objective, which exploits problem structure to allow for scalable optimization.
We develop a novel local combinatorial optimization algorithm that exploits low-rank updates for efficient local search. 
Our framework is both time- and memory-efficient, markedly enhancing the practicality and performance of one-shot structured pruning methods.
For example, on OPT-2.7B, \modelaname~can lead to $125\times$ lower test perpexity on WikiText with $2\times$ inference time speedup in comparison to state-of-the-art ZipLM approach. Our pruning framework takes $6\times$ -- $8\times$ lesser time to prune the network.
Our framework can also prune $100\times$ larger models than previous state-of-the-art structured pruning frameworks.   



\section*{Acknowledgements}
This research is supported in part by grants from the Office of Naval Research (N000142112841 and N000142212665) and Google. We acknowledge the MIT SuperCloud and Lincoln Laboratory Supercomputing Center for providing HPC resources that have contributed to the research results reported within this paper. Additionally, we thank Google for providing us with Google Cloud Credits to run some of the computational experiments reported in this paper. We thank Wenyu Chen and Riade Benbaki for helpful discussions. 

\bibliography{references}
\bibliographystyle{arxiv2023}

\newpage
\appendix
\onecolumn
\section*{Supplement}


\section{Proofs of Main Results}\label{sect:proof}

\subsection{Proof of Proposition \ref{thm:complexity}}\label{sect:prooftc}

We first establish some key notations. For a given matrix $H$ and two subsets of indices $I_1$ and $I_2$, \( H_{I_1,I_2} \) represents the submatrix of \( H \) that includes rows indexed by $I_1$ and columns indexed by $I_2$; $H_{I_1,:}$ represents  the submatrix of \( H \) containing only the rows in $I_1$; $H_{:,I_2}$ represents  the submatrix of \( H \) containing only the columns in $I_2$. Moreover, $\mathbf{E}_a$ represents the identity matrix of size $a\times a$, and $\mathbf{0}_{a\times b}$ refers to the zero matrix of dimensions $a\times b$. For any given set $S\subset [p]$, we denote $t_S=|I_S|$, where $I_S=\left\{i\mid i\in C^j \text{ for some } j \notin S\right\}$.

Let us assume we have two sets \( S \) and \( S' \) with \( t = |I_S\Delta I_{S'}| \). We have already calculated the quadratic coefficients $H\in \mathbb{R}^{d_1\times d_1}$ and linear coefficients $G\in \mathbb{R}^{d_1\times d_2}$ in \eqref{eq:deff}. Additionally, we've computed the value of 
\begin{equation}
    f(S')=  -\frac{1}{2}\Tr\left(G_{I_{S'},:}^\top (H_{I_{S'},I_{S'}})^{-1}G_{I_{S'},:}\right),
\end{equation}
the inverse of \( H_{I_{S'},I_{S'}} \), and the optimal weight matrix for \( f(S') \), denoted as $W^{(S')}$. The remaining part of the proof will demonstrate that the value of \( f(S) \), the inverse of \( H_{I_{S},I_{S}} \) and the optimal weight matrix $W^{(S)}$ for $f(S)$ can be computed with a time complexity of \( O(td_1(d_1+d_2)) \).

Based on the inclusion relationship of sets \(I_S\) and \(I_{S'}\), we consider the following \textit{three} cases.\

\noindent\textbf{Case 1: $I_S\subset I_{S'}$.}

In this case we have $t=|I_r|:=|I_{S'}\backslash I_S|$. Without loss of generality, we can rearrange the rows and columns of $H_{I_{S'},I_{S'}}$ such that $H_{I_{S'},I_{S'}}$ and its inverse are structured as follows:
    \begin{equation}\label{eq:tmp3}
    \renewcommand{\arraystretch}{2}
      H_{I_{S'},I_{S'}} =   \left[ 
\begin{array}{c;{2pt/2pt}c}
        H_{I_S,I_S} & H_{I_S,I_r} \\ \hdashline[2pt/2pt]
        H_{I_r,I_S} & H_{I_r,I_r} 
    \end{array}
\right] ,\,\, \left(H_{I_{S'},I_{S'}}\right)^{-1}=\left[ 
\begin{array}{c;{2pt/2pt}c}
        A & B \\ \hdashline[2pt/2pt]
        B^\top & C
    \end{array}\right],
\end{equation}
where $A$ is a submatrix of $\left(H_{I_{S'},I_{S'}}\right)^{-1}$ with the same size as $H_{I_{S},I_S}$, $B$ has the same size as $H_{I_S,I_r}$, and $C$ has the same size as $H_{I_r,I_r}$. Note that we have computed the $\left(H_{I_{S'},I_{S'}}\right)^{-1}$, so we have the values of $A$, $B$ and $C$.

From the equation
\begin{equation}
    H_{I_{S'},I_{S'}} \left(H_{I_{S'},I_{S'}}\right)^{-1} = \mathbf{E}_{t_{S'}}
\end{equation}
it follows that
\begin{align}
     H_{I_S,I_S}A+H_{I_S,I_r}B^\top &=\mathbf{E}_{t_S} \label{eq:tmp1}\\
     H_{I_S,I_S}B+H_{I_S,I_r}C&=\mathbf{0}_{t_S\times t}. \label{eq:tmp2}
\end{align}
From \eqref{eq:tmp2}, we deduce 
\begin{equation}
 H_{I_S,I_r}=-H_{I_S,I_S}BC^{-1},   
\end{equation}
and substituting this into \eqref{eq:tmp1}  yields
\begin{equation}
 \left(H_{I_S,I_S}\right)^{-1}=A-BC^{-1}B^\top.   
\end{equation}
Given that $B$ is a $t_S\times t$ matrix, and $C$ is a $t\times t$ matrix, computing $=A-BC^{-1}B^\top$ requires $O(t^3+t_S^2t)=O(td_1(d_1+d_2))$ complexity.

Next, we focus on computing $W^{(S)}$.  This is the optimal weight matrix that resolves the following quadratic problem:
\begin{equation}
\begin{aligned}
   \min_{W}\,\,\,& L(W) = \frac{1}{2}\text{Tr}(W^\top H W) - \text{Tr}(G^\top W) \\
       \text{s.t. }\,\,\, & \,\,\wbf_i=0,\,\, \forall\,i\in Q^j,\,j\in S.
\end{aligned}
\end{equation}
This quadratic problem can be solved analytically, yielding the optimal solution $W^{(S)}$ as:
\begin{equation}\label{eq:tmp5}
    W^{(S)}_{I_S,:} = \left(H_{I_S,I_S}\right)^{-1}G_{I_S,:},
\end{equation}
where all elements in $W^{(S)}$ that do not belong to rows in $I_S$ are zero. Similarly, we can express $W^{(S')}$ as 
\begin{equation}
    W^{(S')}_{I_{S'},:} = \left(H_{I_{S'},I_{S'}}\right)^{-1}G_{I_{S'},:}
\end{equation}
where all elements in $W^{(S')}$ that do not belong to rows in $I_{S'}$ are zero. We may further simplify the expression of $W^{(S')}$ as
\begin{equation}\renewcommand{\arraystretch}{2}
    \begin{aligned}
        W^{(S')}_{I_{S'},:} &=  \left[ 
\begin{array}{c}
       W^{(S')}_{I_{S},:} \\ \hdashline[2pt/2pt]
       W^{(S')}_{I_r,:}
    \end{array}\right] = \left[ 
\begin{array}{c;{2pt/2pt}c}
        A & B \\ \hdashline[2pt/2pt]
        B^\top & C
    \end{array}\right]  \left[ 
\begin{array}{c}
        G_{I_{S},:} \\ \hdashline[2pt/2pt]
        G_{I_r,:} 
    \end{array}\right] = \left[ 
\begin{array}{c}
       A G_{I_{S},:} +  BG_{I_r,:}\\ \hdashline[2pt/2pt]
       B^\top G_{I_{S},:} +  CG_{I_r,:}
    \end{array}\right] \\
    &=\left[ 
\begin{array}{c}
       (A-BC^{-1}B^\top) G_{I_{S},:} +  \left(BG_{I_r,:}+BC^{-1}B^\top G_{I_{S},:}\right) \\ \hdashline[2pt/2pt]
       B^\top G_{I_{S},:} +  CG_{I_r,:}
    \end{array}\right]
    =\left[ 
\begin{array}{c}
       W^{(S)}_{I_{S},0}  \\ \hdashline[2pt/2pt]
       \mathbf{0}_{t\times d_2}
    \end{array}\right]+ \left[ 
\begin{array}{c}
        BC^{-1}W^{(S')}_{I_r,:} \\ \hdashline[2pt/2pt]
       W^{(S')}_{I_r,:}
    \end{array}\right]
    \end{aligned}
\end{equation}
This implies that we can compute $W^{(S)}$ as 
\begin{equation}\label{eq:tmp4}\renewcommand{\arraystretch}{2}
    W^{(S)}_{I_{S'},:} = \left[ 
\begin{array}{c}
       W^{(S)}_{I_{S},0}  \\ \hdashline[2pt/2pt]
       \mathbf{0}_{t\times d_2}
    \end{array}\right] = W^{(S')}_{I_{S'},:} - \left[ 
\begin{array}{c}
        BC^{-1} \\ \hdashline[2pt/2pt]
       \mathbf{E}_{t\times d_2}
    \end{array}\right]W^{(S')}_{I_r,:}.
\end{equation}
Given that $B$ is a $t_S\times t$ matrix, $C$ is a $t\times t$ matrix, and $W^{(S')}_{I_r,:}$ is a $t\times d_2$ matrix, computing $W^{S}$ requires $O(t_Std_2)=O(td_1(d_1+d_2))$ time complexity.

Finally, we compute the value of $f(S)$. Given that the optimal weight matrix for $f(S')$ is $W^{(S')}$, we can express 
\begin{equation}
    f(S')=\frac{1}{2}\Tr\left( (W^{(S')})^\top H W^{(S')}\right) - \Tr\left(G^\top W^{(S')}\right).
\end{equation}
It follows a similar argument that
\begin{equation}
     f(S) =\frac{1}{2}\Tr\left( (W^{(S)})^\top H W^{(S)}\right) - \Tr\left(G^\top W^{(S)}\right).
\end{equation}
Together with \eqref{eq:tmp4}, we obtain by some calculations that
\begin{equation}\label{eq:tmp6}
    \begin{aligned}
    f(S')&=\frac{1}{2}\Tr\left( (W^{(S')})^\top H W^{(S')}\right) - \Tr\left(G^\top W^{(S')}\right)\\
    &=\frac{1}{2}\Tr\left( (W^{(S')}_{I_{S'},:})^\top H_{I_{S'},I_{S'}} W^{(S')}_{I_{S'},:}\right) - \Tr\left(G_{I_{S'},:}^\top W^{(S')}_{I_{S'},:}\right)\\
    & = f(S) + \frac{1}{2}\Tr\left( (W^{(S')}_{I_r,:})^\top \renewcommand{\arraystretch}{2}\left[ 
\begin{array}{c;{2pt/2pt}c}
        BC^{-1} & 
       \mathbf{E}_{t\times d_2}
    \end{array}\right] H_{I_{S'},I_{S'}} \left[ 
\begin{array}{c}
        BC^{-1} \\ \hdashline[2pt/2pt]
       \mathbf{E}_{t\times d_2}
    \end{array}\right] W^{(S')}_{I_r,:}\right) \\
    &\quad + \Tr\left((W^{(S')}_{I_r,:})^\top \left[\begin{array}{c;{2pt/2pt}c}
        BC^{-1} & 
       \mathbf{E}_{t\times d_2}
    \end{array}\right] \left(H_{I_{S'},I_{S'}} W^{(S')}_{I_{S'},:}-G_{I_{S'},:}\right)\right)\\
    & = f(S) + \frac{1}{2} \Tr\left( (W^{(S')}_{I_r,:})^\top C^{-1} W^{(S')}_{I_r,:}\right) 
\end{aligned}
\end{equation}
Given that $C$ is a $t\times t$ matrix, and $W^{(S')}_{I_r,:}$ is a $t\times d_2$ matrix, computing $f(S)$ requires $O(t^2d_2)$ time complexity.

We emphasize that, unlike Cases 2 and 3, computing $f(S)$ in this case only requires a time complexity of $O(t^2d_2)$, as opposed to $O(td_1(d_1+d_2))$. This key observation aids in reducing the time cost of Algorithm \ref{alg:localsearch} under certain choices of parameters. We formally present this observation as a lemma:

\begin{lemma}\label{lemma:complexity}
  Given two sets \( S \) and \( S' \) such that $I_S\subset I_{S'}$ and $t=|I_{S'}\backslash I_{S}|$. Suppose we have computed the value of \( f(S') \), the inverse of \( H_{I_{S'},I_{S'}} \) and the optimal weight matrix for \( f(S') \). The value of \( f(S) \) can then be computed within \( O(t^2d_1) \) time complexity.  
\end{lemma}

\noindent\textbf{Case 2: $I_{S'}\subset I_{S}$.}

In this case we have $t=|I_r|:=|I_{S}\backslash I_{S'}|$. Without loss of generality, we can rearrange the rows and columns of $H_{I_{S},I_{S}}$ such that $H_{I_{S},I_{S}}$ and its inverse are structured as follows:
\begin{equation}
    \renewcommand{\arraystretch}{2}
      H_{I_{S},I_{S}} =   \left[ 
\begin{array}{c;{2pt/2pt}c}
        H_{I_{S'},I_{S'}} & H_{I_{S'},I_r} \\ \hdashline[2pt/2pt]
        H_{I_r,I_{S'}} & H_{I_r,I_r} 
    \end{array}
\right] ,\,\, \left(H_{I_{S},I_{S}}\right)^{-1}=\left[ 
\begin{array}{c;{2pt/2pt}c}
        A & B \\ \hdashline[2pt/2pt]
        B^\top & C
    \end{array}\right],
\end{equation}
where $A$ is a submatrix of $\left(H_{I_{S},I_{S}}\right)^{-1}$ with the same size as $H_{I_{S'},I_{S'}}$, $B$ has the same size as $H_{I_{S'},I_r}$, and $C$ has the same size as $H_{I_r,I_r}$. Following a similar argument as in Case 1, we deduce that
\begin{equation}
\begin{aligned}
    C &= \left( H_{I_r,I_r} - H_{I_r,I_{S'}} \left(H_{I_{S'},I_{S'}}\right)^{-1}H_{I_{S'},I_r}  \right)^{-1},\\
    B&= -\left(H_{I_{S'},I_{S'}}\right)^{-1}H_{I_{S'},I_r} C,\\
    A& = \left(H_{I_{S'},I_{S'}}\right)^{-1}+BC^{-1}B.
\end{aligned}
\end{equation}
Given that $\left(H_{I_{S'},I_{S'}}\right)^{-1}$ has already been computed, we can efficiently calculate $A$, $B$, and $C$, and consequently $\left(H_{I_{S},I_{S}}\right)^{-1}$ in $O\left((t_{S'})^2t\right)=O(td_1(d_1+d_2))$ time.

Next, we focus on computing $W^{(S)}$. Using \eqref{eq:tmp5}, we obtain the following relationship between $W^{(S)}$ and $W^{(S')}$:
\begin{equation}\renewcommand{\arraystretch}{2}
    \begin{aligned}
        W^{(S)}_{I_{S},:} &=  \left[ 
\begin{array}{c}
       W^{(S)}_{I_{S'},:} \\ \hdashline[2pt/2pt]
       W^{(S)}_{I_r,:}
    \end{array}\right] = \left[ 
\begin{array}{c;{2pt/2pt}c}
        A & B \\ \hdashline[2pt/2pt]
        B^\top & C
    \end{array}\right]  \left[ 
\begin{array}{c}
        G_{I_{S'},:} \\ \hdashline[2pt/2pt]
        G_{I_r,:} 
    \end{array}\right] = \left[ 
\begin{array}{c}
       A G_{I_{S'},:} +  BG_{I_r,:}\\ \hdashline[2pt/2pt]
       B^\top G_{I_{S'},:} +  CG_{I_r,:}
    \end{array}\right] \\
    &=\left[ 
\begin{array}{c}
       (H_{I_{S'},I_{S'}})^{-1} G_{I_{S'},:} +  \left(BG_{I_r,:}+BC^{-1}B^\top G_{I_{S'},:}\right) \\ \hdashline[2pt/2pt]
       B^\top G_{I_{S'},:} +  CG_{I_r,:}
    \end{array}\right]
    =\left[ 
\begin{array}{c}
       W^{(S')}_{I_{S'},0}  \\ \hdashline[2pt/2pt]
       \mathbf{0}_{t\times d_2}
    \end{array}\right]+ \left[ 
\begin{array}{c}
        BC^{-1} \\ \hdashline[2pt/2pt]
       \mathbf{E}_{t\times d_2}
    \end{array}\right]\left(B^\top G_{I_{S'},:} +  CG_{I_r,:}\right)
    \end{aligned}
\end{equation}
Given that $B$ is a $t_{S'}\times t$ matrix, $C$ is a $t\times t$ matrix, $G_{I_{S'},:}$ is a $(t_{S'})\times d_2$ matrix, and $B,\,C$ can be computed in $O(td_1(d_1+d_2))$ time, computing $W^{(S)}$ requires $O\left((t_{S'})td_2+td_1^2\right)=O(td_1(d_1+d_2))$ time complexity.

Finally, we compute the value of $f(S)$. Using the similar argument as \eqref{eq:tmp6}, we obtain that
\begin{equation}\renewcommand{\arraystretch}{2}
\begin{aligned}
  f(S) &= f(S') + \frac{1}{2} \Tr\left( (W^{(S)}_{I_r,:})^\top C^{-1} W^{(S)}_{I_r,:}\right).
\end{aligned}
\end{equation}
Given that $C$ is a $t\times t$ matrix, and $W^{(S)}_{I_r,:}$ can be computed in $O(td_1(d_1+d_2))$ time, computing $f(S)$ also requires $O(td_1(d_1+d_2))$  time complexity.

\noindent\textbf{Case 3:  no inclusion relationship between $I_{S'}$ and $I_{S}$.}

We define $S''=S'\cup S$, and from the definition of $I_S$, it follows that $I_{S''}=I_S\cap I_{S'}$. Let's denote $t_1=|I_{S'}\backslash I_{S''}|$ and $t_2=|I_{S}\backslash I_{S''}|$. The symmetry difference definition implies that $t=t_1+t_2$.

We have already computed $f(S')$, the inverse of \( H_{I_{S'},I_{S'}} \), and the optimal weight matrix $W^{(S')}$. Note that $I_{S''}\subset I_{S'}$, we can apply the results from Case 1. This allows us to compute the value of $f(S'')$, the inverse of \( H_{I_{S''},I_{S''}} \), and the optimal weight matrix $W^{(S'')}$ in $O(t_1d_1(d_1+d_2))$ time. Moreover, since $I_{S''}\subset I_{S}$, by applying the results from Case 2, we conclude that the value of $f(S)$, the inverse of \( H_{I_{S},I_{S}} \), and the optimal weight matrix $W^{(S)}$ can also be computed within $O(t_2d_1(d_1+d_2))$ time complexity. Therefore, the overall time complexity is $O(td_1(d_1+d_2))$. This completes the proof.

\subsection{Proof of Proposition \ref{thm:alg}}\label{sect:proofalg}

We begin by analyzing the time complexity of Algorithm \ref{alg:localsearch} with general parameters $t_i$ and $p_i$. 
To initiate the algorithm, we need to compute $H^{-1}$ and $H^{-1}G$, which has a time complexity of $O(d_1^2(d_1+d_2))$ since $H$ is a $d_1 \times d_1$ matrix and $G$ is a $d_1 \times d_2$ matrix. 

During iteration $i$ of Algorithm \ref{alg:localsearch}, a local search is conducted by approximately solving \eqref{eq:local} with $S'=S_{i-1}$, $\hat t=t_i$, and $\hat p=p_i$. As outlined in Section \ref{sect:algcomb}, to address \eqref{eq:local}, it suffices to calculate the impact of each element in $[p]$, as defined in \eqref{eq:impact}. Assuming uniform group size $Q$, for any $j\in S_{i-1}$, $|I_{S_{i-1}\backslash \{j\}}\Delta I_{S_{i-1}}|=Q$, and for any $j\notin S_{i-1}$, $|I_{S_{i-1}\cup \{j\}}\Delta I_{S_{i-1}}|=Q$. Having computed $f(S_{i-1})$, the inverse of $H_{I_{S_{i-1}},I_{S_{i-1}}}$, and the optimal weight matrix for $f(S_{i-1})$, Proposition \ref{thm:complexity} suggests that each $B_j$ in \eqref{eq:impact} can be computed in $O(Qd_1(d_1+d_2))$ time. Consequently, computing $B_j$ for all $j\in[p]$ takes $O(d_1^2d_2)$ time, given that $[d_1]$ is partitioned into $p$ equal-sized disjoint groups $Q^1,\,Q^2,\dots,Q^p$ and $pQ=d_1$. 

Upon solving \eqref{eq:local}, we obtain the set $S_i$ with $|S\Delta S'|\le t_i$, which implies $|I_{S_{i}}\Delta I_{S_{i-1}}|\le t_iQ$. Applying Proposition \ref{thm:complexity} again, we can calculate $f(S_i)$, the inverse of \( H_{I_{S_i},I_{S_i}} \), and the optimal weight matrix for \( f(S_i) \) in $O(t_iQd_1(d_1+d_2))=O(d_1^2(d_1+d_2))$ time complexity. Hence, the total time complexity of Algorithm \ref{alg:localsearch} over $T$ iterations is $O(Td_1^2(d_1+d_2))$.

We now focus on the specific parameter choice where $t_i=p_i$ for all $i\in [T]$. During iteration $i$ of Algorithm \ref{alg:localsearch}, the process is similar to the previous case in that we still need to solve \eqref{eq:local}. However, the key difference lies in the fact that since $t_i=p_i$, as outlined in Section \ref{sect:algcomb}, we won't remove elements from $S_{i-1}$. Consequently, it's only necessary to compute $B_j$ for $j\notin S_{i-1}$. This involves calculating $f(S_{i-1}\cup \{j\})-f(S_{i-1})$ for all $i\notin S_{i-1}$. Given that $I_{S_{i-1}\cup \{j\}}\subset I_{S_{i-1}}$ and $Q=|I_{S_{i-1}} \backslash I_{S_{i-1}\cup \{j\}}|$, Lemma \ref{lemma:complexity} implies that $f(S_{i-1}\cup \{j\})-f(S_{i-1})$ can be computed in $O(Q^2d_1)$ time. Thus, computing $B_j$ for all $j\notin S_{i-1}$ requires $O(pQ^2d_1)=O(Qd_1^2)$ time. 

After solving \eqref{eq:local} and obtaining the set $S_i$, similarly, we can apply Proposition \ref{thm:complexity} to compute $f(S_i)$, the inverse of \( H_{I_{S_i},I_{S_i}} \), and the optimal weight matrix for \( f(S_i) \), all within $O(t_iQd_1(d_1+d_2)$ time. The overall time complexity of Algorithm \ref{alg:localsearch} over $T$ iterations is thus 
\begin{equation}
    O\Big(\sum_{i=1}^T  Qd_1^2 + t_iQd_1(d_1+d_2)\Big) = O\left(\Big(\sum_{i=1}^Tt_i\Big)Qd_1(d_1+d_2) \right)
\end{equation} 
Notably, with $t_i=p_i$ for all $i\in [T]$, we obtain $\sum_{i=1}^T t_i=\sum_{i=1}^Tp_i=p'\le \frac{d_1}{Q}$. Therefore, the overall time complexity in this case can be reduced to $O(d_1^2(d_1+d_2))$.

Finally, we turn to the memory complexity of Algorithm \ref{alg:localsearch}. It's important to note that, throughout the algorithm, we only need to maintain the value of $f(S)$, the inverse of \( H_{I_{S},I_{S}} \), and the optimal weight matrix for \( f(S) \) associated with the current solution $S_i$. As shown in Proposition \ref{thm:complexity}, computing these elements involves matrix multiplication with dimensions up to $d_1 \times (d_1 + d_2)$. Consequently, the overall memory complexity of the algorithm is $O(d_1(d_1 + d_2))$.


\section{Experimental Details}

\subsection{Experimental setup}\label{sect:expt-setup}

All experiments were carried out on a computing cluster. Experiments were run on an Intel Xeon Gold 6248 machine with 20 CPUs and a single NVIDIA V100 GPU. Our machine is equipped with 192GB of CPU RAM and 32GB of CUDA memory. We will terminate the algorithm if it leads to Out-of-Memory errors. For our experiments, we use the PyTorch library \cite{paszke2017automatic} to implement all neural network models and pruning methods.

\subsubsection{CNN experiments}\label{sect:cnnsetup}

\noindent\textbf{Pruning settings.}
In our study, we focus exclusively on pruning the convolutional layers within the network. For a layer with $p$ channels, we prune $p'=\tau p$ channels with various $\tau$ values to investigate different speedup ratios. To evaluate the effectiveness of all structured pruning methods, we utilize a calibration dataset consisting of $500$ training samples.

\noindent\textbf{Implementation details.}
Below are the configuration and implementation details for the competing methods and our framework, \modelaname. It's important to note that as we assess the performance of competing methods in a one-shot setting, we shut down the fine-tuning procedure.
\begin{itemize}
    \item \modelaname: We formulate the channel pruning problem as a combinatorial problem (see \eqref{eq:local}), and address it using Algorithm \ref{alg:localsearch}. We set the number of iterations $T=p'/2$ and $t_i=p_i=2$ for $i\in[T]$.
    \item MP~\cite{mozer1989using,He2018a}: We implement structured pruning by calculating the Frobenius norm of weights for each channel as its magnitude and pruning $p'$ channels with the smallest magnitude.
    \item CHIP~\cite{Sui2021}: We utilize the authors' algorithm (codes available on \href{https://github.com/Eclipsess/CHIP_NeurIPS2021}{GitHub}) to prune channels based on channel independence.
    \item FPGM~\cite{He2018geometric}: We utilize the authors' algorithm (codes available on \href{https://github.com/he-y/filter-pruning-geometric-median}{GitHub}) for channel pruning via the geometric median.
    \item Lasso~\cite{He2017}: We utilize our own implementation, as directly applying the authors' codes to our setting is challenging. Following \citep[Section~3]{He2017}, we convert the channel pruning problem into a $\ell_1$-regularized problem. We solve the subproblem related to $\beta$ repeatedly with increasing $\ell_1$-penalty coefficients until meeting the pruning constraint, then address the subproblem related to the weight matrix $W$. Our implementation of this algorithm strictly adheres to the description provided by the authors in their paper.
    \item ThiNet~\cite{Luo17}:  We utilize the authors' algorithm (codes available on \href{https://github.com/Roll920/ThiNet_Code}{GitHub}) 
     to conduct channel selection through a greedy approach, followed by a refining process to minimize reconstruction error.
\end{itemize}

\noindent\textbf{Pruning strategies.}
Additionally, to further enhance accuracy, we implement two techniques for all \modelaname~and all competing methods. Firstly,  as advised by \cite{zhang2015accelerating} and \citep[Section~3.2]{He2017}, we solve the channel pruning problem layer by layer sequentially. For each layer, we minimize the squared loss between the output of the \textit{dense} model at this layer and the output produced by the pruned weight \( W \) acting on the pruned model's input feature map. Thus, our layer-wise pruning objective for the $\ell$-th layer is given as:
\begin{equation}
  L(w) =  \frac{1}{N}\sum_{i=1}^N \left\|\Conv\left(\widehat{w},\widehat{X}^{i}\right)-\Conv\left(w,X^{i}\right)\right\|^2,
\end{equation}
Different from \eqref{eq:obj-ver1}, here we replace $\Conv\left(\widehat{w},X^{i}\right)$ by $\Conv\left(\widehat{w},\widehat{X}^{i}\right)$, where $\widehat{X}$ denotes the dense model's input feature map, while $X$ represents the pruned model's input feature map (with the first $\ell-1$ layers pruned). This modification does not change the nature of the problem, and it enables pruning algorithms to more accurately approximate the output of the original model.

Secondly, for ResNet20 and ResNet50, following insights from \citep[Section~3.3]{Luo17}, we avoid pruning the first convolution layer in each residual block. This approach keeps block output dimensions unchanged and avoids extra costs in executing the residual connection. For MobileNetV1,  we focus the channel pruning on the $1\times1$ convolutional layers as pruning the depthwise convolution layers tends to be less impactful and could significantly reduce accuracy.  It's crucial to highlight that layers not directly pruned in the process are still non-dense in the pruned model. This is because some filters in these layers will become redundant when the channels they contribute to in subsequent layers are pruned. Hence, these filters can be pruned too.
 
\subsubsection{LLM experiments}\label{sect:llmsetup}

\noindent\textbf{Pruning settings.}

As discussed in Section \ref{sect:problem-formulation}, for pruning OPT models, we focus on removing attention heads and reducing the intermediate dimension in fully connected layers. We convert these two problems into MIQP problems \eqref{eq:MIQP}, and we set $p'=\tau p$ channels with various $\tau$ values to investigate different speedup ratios. Here $p$ is either the number of attention heads or the input dimension of the fully connected layer, depending on the sublayer we are pruning.

\noindent\textbf{Implementation details.}
Below are the configuration and implementation details for the competing methods and our framework, \modelaname. 
\begin{itemize}
    \item \modelaname: We utilize Algorithm \ref{alg:localsearch} to solve the combinatorial problem derived from structured pruning. For pruning attention heads, we set the number of iterations $T=p'/2$ and $t_i=p_i=2$ for $i\in[T]$.  For reducing intermediate dimensions in fully connected layers, we set the number of iterations $T=p'/10$ and $t_i=p_i=10$ for $i\in[T]$.
    \item MP~\cite{mozer1989using}: This method employs structured pruning by calculating the Frobenius norm of each weight group $Q^i,\,i=1,2,\dots,p$ (as defined in \eqref{eq:MIQP}) as its magnitude and then pruning $p'$ groups with the lowest magnitude.
    \item MP+: This approach extends MP by further refining the weights post-pruning. After performing MP, the weight matrix $W$ is updated following \eqref{eq:deff} to reduce the approximation loss $L(W)$, where the set $S$ is determined by MP. 
    \item ZipLM \cite{ziplm}: ZipLM's original implementation targeted models smaller than 340 million parameters. We adapt their framework for OPT models. In ZipLM's second stage, a structured SPDY search is performed, involving a 1000-step neighborhood search for various sparsity configurations across layers. Each configuration's performance is assessed on a validation set, with the best-performing configuration in terms of perplexity being selected. However, for larger OPT models, we found this SPDY search to be computationally prohibitive. For example, based on our estimation, it would take $\sim 200$ days for ZipLM to perform 100 search steps and validate each configuration's performance for OPT-2.7B. Consequently, we turn off the SPDY search, using on the original layer-wise sensitivity coefficients to generate the sparsity configuration.
\end{itemize}

\noindent\textbf{Pruning strategies.} 
Similar to the strategy used in CNNs, we solve the structured pruning problem layer by layer sequentially in OPT models. For each layer, our objective is to minimize the squared loss between the output of the dense model at that particular layer and the output generated by the pruned weight \( W \) acting on the pruned model's input values. It's important to note that for LLMs, this strategy is implemented only for \modelaname, MP, and MP+, and not for ZipLM. In the case of ZipLM, as outlined in their paper, involves pruning each layer to different sparsity levels.  This requires considering the squared loss between the output from the dense weight \( \widehat{W} \) acting on the dense model's input values and the output from the pruned weight \( W \) acting on the dense model's input values.

\noindent\textbf{Saving GPU memory.}  
We employ a layer-by-layer pruning approach for \modelaname~and all competing methods. Consequently, we only need to load the weights of a single layer into GPU memory at any given time, offloading other weights to the CPU to conserve GPU memory. Additionally, for our calibration data – comprising 128 segments of 2048 tokens each from the C4 dataset – we store these segments on the CPU to further save GPU memory. They are loaded to the GPU individually when we need them for generating $H$ and $G$ in the pruning objective \eqref{eq:obj-ver2}. This strategy significantly reduces GPU memory usage, allowing us to apply \modelaname~for pruning large OPT models with just 32GB of GPU memory.


\subsection{Ablation studies and additional results}\label{sect:addexp}
In this section, we provide additional experimental results and ablation studies focusing on one-shot structured pruning in both CNNs and LLMs.

\subsubsection{Pruning performance on a single convolutional layer.}
\begin{figure}[!t]
     \centering
     \includegraphics[width=0.6\columnwidth]{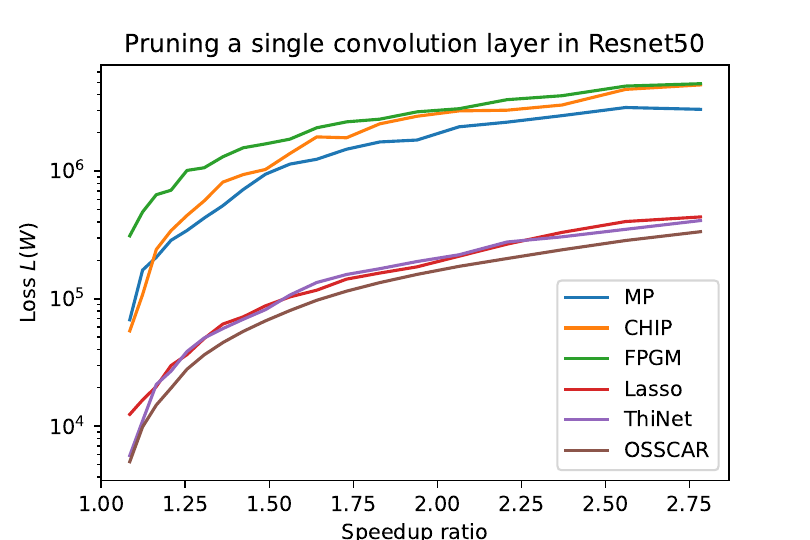}
     \caption{Performance of pruning the layer ``layer1.0.conv2" in Resnet50 at different speedup ratios. The plot shows the squared loss between dense and pruned weights of this single layer, using different pruning methods. The speedup ratio is computed as the total number of channels (64) divided by the number of channels retained after pruning.
     }
     \label{fig:single}
\end{figure}

We first examine structured pruning on a single convolutional layer in ResNet50. 
Figure \ref{fig:single} displays the squared loss between dense weights and pruned weights at different speedup ratios. Our method \modelaname~outperforms MP, CHIP, and FPGM significantly, as these methods do not focus on directly minimizing \( L(W) \). When compared to ThiNet and Lasso, which do aim to minimize $L(W)$, \modelaname~still achieves about 20\% lower loss under the same speedup ratios. These results highlight the effectiveness of our proposed local optimization method, as outlined in Algorithm \ref{alg:localsearch}, in obtaining a high-quality solution for the layer-wise channel pruning problem.

\subsubsection{Pruning performance of OPT models on PTB and C4 datasets}
\label{sect:results-on-ptb-c4}
In this section, we present the perplexity performance of MP, MP+, ZipLM, and \modelaname~on the PTB (test) and C4 (validation) datasets, specifically for pruning OPT-1.3B, OPT-2.7B, and OPT-6.7B models. The results are detailed in Tables \ref{tab:osscar-llm-ptb} and \ref{tab:osscar-llm-c4}. Similar to the trends observed with the Wikitext dataset, \modelaname~demonstrates a significant reduction in perplexity compared to magnitude pruning based methods and ZipLM (state-of-the-art), particularly at higher speedup ratios.

\begin{table}[!h]
\centering
\captionsetup{width=0.75\textwidth}
\caption{Perplexity performance on PTB for one-shot structured pruning of OPT models (1.3B, 2.7B, and 6.7B). The speedup ratio denotes the inference time improvement of pruned models over dense models. For all methods we take ten runs and report the mean and standard error.}
\label{tab:osscar-llm-ptb}
\setlength{\tabcolsep}{15pt}
\resizebox{0.75\textwidth}{!}{%
\begin{tabular}{c|c|c|c|c|c}
\toprule
Model & Speedup & MP & MP+ & ZipLM & \modelaname \\ \midrule
\multirow{7}{*}{OPT-1.3B} & 1.2x & 164.8 ($\pm$0.06) & 24.54 ($\pm$0.08) & \textbf{18.57} ($\pm$0.21) & 19.79 ($\pm$0.14) \\
 & 1.3x & 1166 ($\pm$0.72) & 48.60 ($\pm$0.83) & 63.61 ($\pm$2.63) & \textbf{24.45} ($\pm$0.23) \\
 & 1.4x & 7147 ($\pm$20.8) & 795.0 ($\pm$53.1) & 477.9 ($\pm$62.0) & \textbf{30.08} ($\pm$0.42) \\
 & 1.7x & 7762 ($\pm$31.8) & 1646 ($\pm$90.8) & 1786 ($\pm$161) & \textbf{40.74} ($\pm$1.39) \\
 & 2.0x & 6796 ($\pm$8.87) & 4320 ($\pm$224) & 3716 ($\pm$386) & \textbf{62.02} ($\pm$2.94) \\
 & 2.6x & 8092 ($\pm$5.89) & 5628 ($\pm$124) & 5330 ($\pm$312) & \textbf{97.92} ($\pm$4.17) \\
 & 3.3x & 9612 ($\pm$30.8) & 6071 ($\pm$127) & 9616 ($\pm$692) & \textbf{170.9} ($\pm$9.65) \\ \midrule
\multirow{7}{*}{OPT-2.7B} & 1.2x & 365.9 ($\pm$1.08) & 24.90 ($\pm$0.09) & \textbf{15.15} ($\pm$0.03) & 17.21 ($\pm$0.20) \\
 & 1.3x & 3281 ($\pm$123) & 38.64 ($\pm$0.23) & 26.91 ($\pm$2.29) & \textbf{19.80} ($\pm$0.30) \\
 & 1.4x & 10062 ($\pm$969) & 88.62 ($\pm$1.14) & 257.5 ($\pm$58.4) & \textbf{24.91} ($\pm$0.41) \\
 & 1.7x & 7053 ($\pm$52.7) & 1380 ($\pm$152) & 1289 ($\pm$176) & \textbf{31.20} ($\pm$0.79) \\
 & 2.0x & 8530 ($\pm$29.8) & 4803 ($\pm$248) & 3267 ($\pm$374) & \textbf{43.47} ($\pm$1.11) \\
 & 2.4x & 6854 ($\pm$16.1) & 7407 ($\pm$166) & 8017 ($\pm$852) & \textbf{70.95} ($\pm$1.81) \\
 & 3.0x & 10166 ($\pm$37.9) & 7780 ($\pm$171) & 12624 ($\pm$968) & \textbf{122.7} ($\pm$3.08) \\ \midrule
\multirow{7}{*}{OPT-6.7B} & 1.1x & 208.2 ($\pm$0.40) & 22.45 ($\pm$0.10) & \multirow{7}{*}{OOM} & \textbf{17.21} ($\pm$0.44) \\
 & 1.2x & 4086 ($\pm$37.8) & 53.57 ($\pm$1.64) &  & \textbf{18.96} ($\pm$0.38) \\
 & 1.5x & 15915 ($\pm$186) & 1056 ($\pm$145) &  & \textbf{22.24} ($\pm$0.58) \\
 & 1.7x & 15668 ($\pm$110) & 2858 ($\pm$162) &  & \textbf{26.78} ($\pm$0.57) \\
 & 2.0x & 11537 ($\pm$71.3) & 5833 ($\pm$90.1) &  & \textbf{34.80} ($\pm$1.10) \\
 & 2.6x & 12524 ($\pm$69.0) & 7142 ($\pm$177) &  & \textbf{54.70} ($\pm$1.20) \\
 & 3.3x & 13585 ($\pm$87.9) & 6815 ($\pm$220) &  & \textbf{102.4} ($\pm$3.57) \\ \bottomrule
\end{tabular}%
}
\end{table}

\begin{table}[!h]
\centering
\captionsetup{width=0.75\textwidth}
\caption{Perplexity performance on C4 for one-shot structured pruning of OPT models (1.3B, 2.7B, and 6.7B). The speedup ratio denotes the inference time improvement of pruned models over dense models. For all methods we take ten runs and report the mean and standard error.}
\label{tab:osscar-llm-c4}
\resizebox{0.75\textwidth}{!}{%
\setlength{\tabcolsep}{15pt}
\begin{tabular}{c|c|c|c|c|c}
\toprule
Model & Speedup & MP & MP+ & ZipLM & \modelaname \\ \midrule
\multirow{7}{*}{OPT-1.3B} & 1.2x & 84.29 ($\pm$0.03) & 18.32 ($\pm$0.04) & \textbf{14.59} ($\pm$0.03) & 15.58 ($\pm$0.04) \\
 & 1.3x & 814.6 ($\pm$0.80) & 26.24 ($\pm$0.21) & 41.81 ($\pm$1.57) & \textbf{16.78} ($\pm$0.09) \\
 & 1.4x & 5630 ($\pm$9.11) & 322.1 ($\pm$15.4) & 332.6 ($\pm$37.7) & \textbf{18.73} ($\pm$0.17) \\
 & 1.7x & 6188 ($\pm$24.6) & 847.5 ($\pm$74.7) & 943.4 ($\pm$39.5) & \textbf{22.02} ($\pm$0.25) \\
 & 2.0x & 6327 ($\pm$8.17) & 2994 ($\pm$46.9) & 1823 ($\pm$103) & \textbf{28.60} ($\pm$0.42) \\
 & 2.6x & 8046 ($\pm$9.65) & 4264 ($\pm$92.3) & 2693 ($\pm$100) & \textbf{46.29} ($\pm$0.78) \\
 & 3.3x & 9611 ($\pm$21.9) & 5365 ($\pm$99.5) & 5643 ($\pm$425) & \textbf{84.49} ($\pm$1.97) \\ \midrule
\multirow{7}{*}{OPT-2.7B} & 1.2x & 296.0 ($\pm$0.73) & 19.62 ($\pm$0.10) & \textbf{12.63} ($\pm$0.00) & 13.85 ($\pm$0.04) \\
 & 1.3x & 4108 ($\pm$164) & 27.94 ($\pm$0.30) & 17.61 ($\pm$0.73) & \textbf{14.88} ($\pm$0.08) \\
 & 1.4x & 11538 ($\pm$1386) & 46.83 ($\pm$0.85) & 189.3 ($\pm$37.4) & \textbf{16.47} ($\pm$0.13) \\
 & 1.7x & 10179 ($\pm$87.7) & 821.7 ($\pm$128) & 712.8 ($\pm$103) & \textbf{19.17} ($\pm$0.23) \\
 & 2.0x & 10345 ($\pm$21.5) & 4298 ($\pm$213) & 1711 ($\pm$195) & \textbf{24.25} ($\pm$0.34) \\
 & 2.4x & 9363 ($\pm$17.3) & 6944 ($\pm$167) & 3660 ($\pm$331) & \textbf{36.91} ($\pm$0.78) \\
 & 3.0x & 14822 ($\pm$40.7) & 7329 ($\pm$111) & 6567 ($\pm$599) & \textbf{65.42} ($\pm$1.17) \\ \midrule
\multirow{7}{*}{OPT-6.7B} & 1.1x & 226.3 ($\pm$0.04) & 16.42 ($\pm$0.10) & \multirow{7}{*}{OOM} & \textbf{12.26} ($\pm$0.04) \\
 & 1.2x & 3694 ($\pm$4.38) & 24.89 ($\pm$0.41) &  & \textbf{13.16} ($\pm$0.10) \\
 & 1.5x & 19913 ($\pm$142) & 271.5 ($\pm$16.9) &  & \textbf{14.48} ($\pm$0.17) \\
 & 1.7x & 17843 ($\pm$19.5) & 1665 ($\pm$122) &  & \textbf{16.56} ($\pm$0.30) \\
 & 2.0x & 16617 ($\pm$51.8) & 5834 ($\pm$264) &  & \textbf{20.31} ($\pm$0.43) \\
 & 2.6x & 18948 ($\pm$41.6) & 7744 ($\pm$203) &  & \textbf{29.55} ($\pm$0.17) \\
 & 3.3x & 20887 ($\pm$56.0) & 8000 ($\pm$126) &  & \textbf{49.70} ($\pm$1.41) \\ \bottomrule
\end{tabular}%
}
\end{table}



\subsubsection{Effect of calibration dataset size}\label{sect:llmsize}

In this study, we explore how varying the sample size of the calibration dataset affects the performance of one-shot structured pruning on LLMs. For all our LLM experiments previously conducted, we utilized a calibration dataset comprising $N=128$ sequences of $2048$ tokens each, sampled from C4. We now experiment with different values of $N$ from the set $\{128,256,512,1024,2048,4096\}$ to assess its impact on the perplexity of pruned models. The results of applying \modelaname~to prune OPT models ranging from 1.3B to 30B for a 2x speedup with these varied sample sizes $N$ are shown in Table \ref{tab:osscar-llm-samples}. It is observed that for a 2x speedup, the performance improves by up to $\sim3.5$ perplexity units when increasing the sample size $N$ from $128$ to $2048$. Beyond this point, the performance plateaus, showing no significant change with further increases in sample size.

\begin{table}[!h]
\small
\centering
\captionsetup{width=\textwidth}
\caption{
Perplexity performance on Wikitext for applying one-shot structured pruning of OPT models (1.3B, 2.7B, 6.7B, 13B and 30B),  with varying sizes of calibration data. 
All pruned models shown in this table have a 2x improvement in inference time compared to dense models. For all models and calibration data sizes, we take ten runs and report the mean and standard error.
}
\label{tab:osscar-llm-samples}
\setlength{\tabcolsep}{15pt}
\resizebox{\textwidth}{!}{%
\begin{tabular}{c|ccccc}
\toprule
\multicolumn{1}{c}{Number of Samples} & \multicolumn{1}{c}{OPT-1.3B} & \multicolumn{1}{c}{OPT-2.7B} & \multicolumn{1}{c}{OPT-6.7B} & \multicolumn{1}{c}{OPT-13B} & OPT-30B \\ \midrule
\multicolumn{1}{c|}{128} & \multicolumn{1}{c}{38.16 ($\pm$0.52)} & \multicolumn{1}{c}{29.49 ($\pm$0.53)} & \multicolumn{1}{c}{25.03 ($\pm$0.73)} & \multicolumn{1}{c}{22.59 ($\pm$0.59)} & 17.11 ($\pm$0.30) \\ 
\multicolumn{1}{c|}{256} & \multicolumn{1}{c}{37.29 ($\pm$0.53)} & \multicolumn{1}{c}{28.29 ($\pm$0.36)} & \multicolumn{1}{c}{23.92 ($\pm$0.21)} & \multicolumn{1}{c}{22.34 ($\pm$0.34)} & 16.45 ($\pm$0.17) \\ 
\multicolumn{1}{c|}{512} & \multicolumn{1}{c}{36.04 ($\pm$0.25)} & \multicolumn{1}{c}{27.91 ($\pm$0.25)} & \multicolumn{1}{c}{23.25 ($\pm$0.25)} & \multicolumn{1}{c}{22.27 ($\pm$0.35)} & 16.07 ($\pm$0.11) \\ 
\multicolumn{1}{c|}{1024} & \multicolumn{1}{c}{35.91 ($\pm$0.09)} & \multicolumn{1}{c}{27.73 ($\pm$0.34)} & \multicolumn{1}{c}{23.76 ($\pm$0.28)} & \multicolumn{1}{c}{21.79 ($\pm$0.29)} & 15.86 ($\pm$0.06) \\ 
\multicolumn{1}{c|}{2048} & \multicolumn{1}{c}{\textbf{34.87} ($\pm$0.11)} & \multicolumn{1}{c}{\textbf{27.09} ($\pm$0.12)} & \multicolumn{1}{c}{23.15 ($\pm$0.15)} & \multicolumn{1}{c}{21.34 ($\pm$0.10)} & \textbf{15.77} ($\pm$0.04) \\ 
\multicolumn{1}{c|}{4096} & \multicolumn{1}{c}{34.97 ($\pm$0.11)} & \multicolumn{1}{c}{27.38 ($\pm$0.30)} & \multicolumn{1}{c}{\textbf{22.91} ($\pm$0.17)} & \multicolumn{1}{c}{\textbf{20.95} ($\pm$0.17)} & OOM* \\ \bottomrule
\multicolumn{6}{l}{${}^{*}$OOM denotes out of CPU memory here.} \\
\end{tabular}%
}
\end{table}

As the number of samples increases, the total time required for applying \modelaname~to prune the model also rises. The pruning process consists of two parts: constructing matrices $H$ and $G$ as defined in \eqref{eq:obj-ver2} for each layer, and solving the structured pruning problem using Algorithm \ref{alg:localsearch}. The time taken to construct matrices $H$ and $G$ grows linearly with the number of samples. However, the time required to apply Algorithm \ref{alg:localsearch} is largely unaffected by the number of samples. This is because the dimensions of $H$ and $G$ ($d_1$ and $d_2$) do not depend on $N$; hence, as per Proposition \ref{thm:alg}, the time complexity of the algorithm remains the same. For instance, when pruning the OPT-13B model with $N=128$ samples, constructing $H$ and $G$ takes 20 minutes, and Algorithm \ref{alg:localsearch} takes 25 minutes; for $N=2048$ samples, constructing $H$ and $G$ takes 320 minutes, while Algorithm \ref{alg:localsearch} takes 28 minutes.


 \begin{table}[!h]
     \centering
      \caption{Test accuracy for applying \modelaname~to prune CNNs (ResNet20, MobileNetV1 and ResNet50), under different choices of parameters $\hat{t}=\hat{p}$. 
 For all models and parameters, we take ten runs and report the mean and standard error.}
        \label{tab:osscar-cnn-hatt}
        \setlength{\tabcolsep}{10pt}
        \resizebox{0.5\textwidth}{!}{%
        \begin{tabular}{c|c|ccc}
        \toprule
        \multicolumn{1}{c}{\multirow{2}{*}{Speedup}} & \multicolumn{1}{c}{\multirow{2}{*}{$\hat{t}=\hat{p}$}} & \multicolumn{3}{c}{Models} \\
        \multicolumn{1}{c|}{} & \multicolumn{1}{c|}{} & \multicolumn{1}{c}{ResNet20} & \multicolumn{1}{c}{MobileNetV1} & ResNet50 \\ \midrule
        \multicolumn{1}{c|}{\multirow{4}{*}{1.4x}} & 1 &  \multicolumn{1}{c}{87.69 (±0.14)} &  \multicolumn{1}{c}{\textbf{68.89} (±0.09)} & 65.29 (±0.29)\\ 
        \multicolumn{1}{c|}{} & 2 &  \multicolumn{1}{c}{\textbf{87.70} (±0.16)} &  \multicolumn{1}{c}{68.89 (±0.08)} & \textbf{65.32} (±0.27)\\ 
        \multicolumn{1}{c|}{} & 5 &  \multicolumn{1}{c}{87.60 (±0.13)} &  \multicolumn{1}{c}{68.88 (±0.10)} & 65.24 (±0.24)\\ 
        \multicolumn{1}{c|}{} & 10 &  \multicolumn{1}{c}{87.51 (±0.11)} &  \multicolumn{1}{c}{68.79 (±0.09)} & 65.23 (±0.23)\\ 
         \midrule
        \multicolumn{1}{c|}{\multirow{4}{*}{1.9-2.0x}} & 1 &  \multicolumn{1}{c}{\textbf{81.34} (±0.50)} &  \multicolumn{1}{c}{\textbf{58.42} (±0.41)} & \textbf{38.56} (±0.64)\\
        \multicolumn{1}{c|}{} & 2&  \multicolumn{1}{c}{81.10 (±0.49)} &  \multicolumn{1}{c}{58.39 (±0.36)} & 38.45 (±0.63)\\
        \multicolumn{1}{c|}{} & 5 &  \multicolumn{1}{c}{80.89 (±0.42)} &  \multicolumn{1}{c}{58.32 (±0.30)} & 38.44 (±0.64)\\
        \multicolumn{1}{c|}{} & 10 &  \multicolumn{1}{c}{80.56 (±0.52)} &  \multicolumn{1}{c}{58.20 (±0.41)} & 38.23 (±0.36)\\ \bottomrule
        \end{tabular}%
        }
 \end{table}

\begin{table}[!h]
\centering
    \begin{minipage}[b]{0.48\textwidth}\centering
        \centering
        \captionsetup{width=\textwidth}
        \caption{ Perplexity performance on Wikitext for applying \modelaname~to prune OPT models (1.3B, 2.7B and 6.7B), under different choices of parameters $\hat{t}=\hat{p}$. 
 For all models and parameters, we take ten runs and report the mean and standard error.
        }
        \label{tab:osscar-llm-t_hat-perpexity}
        \setlength{\tabcolsep}{10pt}
        \resizebox{\textwidth}{!}{%
        \begin{tabular}{c|c|ccc}
        \toprule
        \multicolumn{1}{c}{\multirow{2}{*}{Speedup}} & \multicolumn{1}{c}{\multirow{2}{*}{$\hat{t}=\hat{p}$}} & \multicolumn{3}{c}{Models} \\
        \multicolumn{1}{c|}{} & \multicolumn{1}{c|}{} & \multicolumn{1}{c}{OPT-1.3B} & \multicolumn{1}{c}{OPT-2.7B} & OPT-6.7B \\ \midrule
        \multicolumn{1}{c|}{\multirow{6}{*}{1.4x-1.5x}} & 1 & \multicolumn{1}{c}{20.53 ($\pm$0.35)} & \multicolumn{1}{c}{17.23 ($\pm$0.19)} & 14.94 ($\pm$0.19) \\
        \multicolumn{1}{c|}{} & 5 & \multicolumn{1}{c}{20.54 ($\pm$0.38)} & \multicolumn{1}{c}{\textbf{17.20} ($\pm$0.20)} & 14.93 ($\pm$0.20) \\
        \multicolumn{1}{c|}{} & 10 & \multicolumn{1}{c}{\textbf{20.50} ($\pm$0.36)} & \multicolumn{1}{c}{17.23 ($\pm$0.20)} & 14.93 ($\pm$0.18) \\
        \multicolumn{1}{c|}{} & 50 & \multicolumn{1}{c}{20.60 ($\pm$0.39)} & \multicolumn{1}{c}{17.26 ($\pm$0.19)} & \textbf{14.84} ($\pm$0.20) \\
        \multicolumn{1}{c|}{} & 100 & \multicolumn{1}{c}{20.65 ($\pm$0.37)} & \multicolumn{1}{c}{17.28 ($\pm$0.20)} & 14.89 ($\pm$0.20) \\
        \multicolumn{1}{c|}{} & 500 & \multicolumn{1}{c}{20.65 ($\pm$0.37)} & \multicolumn{1}{c}{17.54 ($\pm$0.22)} & 15.29 ($\pm$0.21) \\ \midrule
        \multicolumn{1}{c|}{\multirow{6}{*}{2.0x}} & 1 & \multicolumn{1}{c}{38.04 ($\pm$0.59)} & \multicolumn{1}{c}{29.58 ($\pm$0.47)} & 24.92 ($\pm$0.79) \\
        \multicolumn{1}{c|}{} & 5 & \multicolumn{1}{c}{38.08 ($\pm$0.55)} & \multicolumn{1}{c}{29.64 ($\pm$0.50)} & 25.00 ($\pm$0.76) \\
        \multicolumn{1}{c|}{} & 10 & \multicolumn{1}{c}{38.16 ($\pm$0.52)} & \multicolumn{1}{c}{\textbf{29.49} ($\pm$0.53)} & 25.02 ($\pm$0.73) \\
        \multicolumn{1}{c|}{} & 50 & \multicolumn{1}{c}{38.08 ($\pm$0.52)} & \multicolumn{1}{c}{29.63 ($\pm$0.52)} & 25.37 ($\pm$0.63) \\
        \multicolumn{1}{c|}{} & 100 & \multicolumn{1}{c}{\textbf{38.03} ($\pm$0.51)} & \multicolumn{1}{c}{29.62 ($\pm$0.58)} & 25.43 ($\pm$0.63) \\
        \multicolumn{1}{c|}{} & 500 & \multicolumn{1}{c}{38.65 ($\pm$0.60)} & \multicolumn{1}{c}{30.04 ($\pm$0.70)} & 25.76 ($\pm$0.63) \\ \bottomrule
        \end{tabular}%
        }
    \end{minipage}
    \hfill
    \begin{minipage}[b]{0.48\textwidth}\centering
        \centering
        \captionsetup{width=\textwidth}
        \caption{Pruning time for applying \modelaname~to prune OPT models (1.3B, 2.7B and 6.7B), under different choices of parameters $\hat{t}=\hat{p}$. 
 For all models and parameters, we take ten runs and report the mean and standard error.}
        \label{tab:osscar-llm-t_hat-time}
        \setlength{\tabcolsep}{14pt}
        \resizebox{\textwidth}{!}{%
        \begin{tabular}{c|c|ccc}
        \toprule
        \multicolumn{1}{c}{\multirow{2}{*}{Speedup}} & \multicolumn{1}{c}{\multirow{2}{*}{$\hat{t}=\hat{p}$}} & \multicolumn{3}{c}{Models} \\
        \multicolumn{1}{c|}{} & \multicolumn{1}{c|}{} & \multicolumn{1}{c}{OPT-1.3B} & \multicolumn{1}{c}{OPT-2.7B} & OPT-6.7B \\ \midrule
        \multicolumn{1}{c|}{\multirow{6}{*}{1.4x-1.5x}} & 1 & \multicolumn{1}{c}{376 ($\pm$8.0)} & \multicolumn{1}{c}{724 ($\pm$5.0)} & 2592 ($\pm$10) \\
        \multicolumn{1}{c|}{} & 5 & \multicolumn{1}{c}{244 ($\pm$21)} & \multicolumn{1}{c}{434 ($\pm$3.0)} & 1140 ($\pm$6.0) \\
        \multicolumn{1}{c|}{} & 10 & \multicolumn{1}{c}{201 ($\pm$6.0)} & \multicolumn{1}{c}{384 ($\pm$3.0)} & 974 ($\pm$4.0) \\
        \multicolumn{1}{c|}{} & 50 & \multicolumn{1}{c}{185 ($\pm$4.0)} & \multicolumn{1}{c}{353 ($\pm$5.0)} & 825 ($\pm$5.0) \\
        \multicolumn{1}{c|}{} & 100 & \multicolumn{1}{c}{185 ($\pm$3.0)} & \multicolumn{1}{c}{348 ($\pm$4.0)} & 805 ($\pm$6.0) \\
        \multicolumn{1}{c|}{} & 500 & \multicolumn{1}{c}{183 ($\pm$5.0)} & \multicolumn{1}{c}{347 ($\pm$6.0)} & 802 ($\pm$6.0) \\ \midrule
        \multicolumn{1}{c|}{\multirow{6}{*}{2.0x}} & 1 & \multicolumn{1}{c}{526 ($\pm$8.0)} & \multicolumn{1}{c}{1108 ($\pm$6.0)} & 4009 ($\pm$13) \\
        \multicolumn{1}{c|}{} & 5 & \multicolumn{1}{c}{256 ($\pm$9.0)} & \multicolumn{1}{c}{500 ($\pm$3.0)} & 1419 ($\pm$4.0) \\
        \multicolumn{1}{c|}{} & 10 & \multicolumn{1}{c}{218 ($\pm$6.0)} & \multicolumn{1}{c}{418 ($\pm$2.0)} & 1118 ($\pm$15) \\
        \multicolumn{1}{c|}{} & 50 & \multicolumn{1}{c}{189 ($\pm$4.0)} & \multicolumn{1}{c}{366 ($\pm$4.0)} & 848 ($\pm$2.0) \\
        \multicolumn{1}{c|}{} & 100 & \multicolumn{1}{c}{189 ($\pm$7.0)} & \multicolumn{1}{c}{361 ($\pm$2.0)} & 821 ($\pm$3.0) \\
        \multicolumn{1}{c|}{} & 500 & \multicolumn{1}{c}{189 ($\pm$3.0)} & \multicolumn{1}{c}{369 ($\pm$11)} & 798 ($\pm$3.0) \\ \bottomrule
        \end{tabular}%
        }
    \end{minipage}%
\end{table}

\subsubsection{Performance of Algorithm \ref{alg:localsearch} across various parameter settings}\label{sect:choicetk}

In this section, we explore how parameters \(\hat t\) and \(\hat p\) in \eqref{eq:local} influence the performance of Algorithm \ref{alg:localsearch}.

We begin by setting $\hat t=\hat p$ to assess the impact of $\hat t$ on both the solution quality and the time complexity of the algorithm. Specifically, we set $t_i=p_i=\hat t$ for all iterations $i\in[T]$ and set the number of iterations $T=p'/\hat t$ for Algorithm \ref{alg:localsearch}

We detail in Table \ref{tab:osscar-cnn-hatt} the accuracy performance of \modelaname~for pruning CNNs with different $\hat{t}$ values. For OPT models, we display the perplexity performance and pruning time in Tables \ref{tab:osscar-llm-t_hat-perpexity} and \ref{tab:osscar-llm-t_hat-time}, respectively. Notably, when pruning attention heads in OPT models, we consistently set $\hat t=2$---in this context, Algorithm \ref{alg:localsearch} typically converges in less than 50 iterations regardless of parameter choice, and varying $\hat t$ offers no advantage. Different $\hat{t}$ values are considered when reducing the intermediate dimension in fully connected layers.

We observe that smaller $\hat{t}$ values yield marginally better performance, although the difference is not statistically significant. This suggests that \modelaname’s accuracy/perplexity performance exhibits low sensitivity to the choice of $\hat{t}$. In contrast, the pruning time can be significantly reduced (2-5x) when increasing $\hat t$ from 1 to 100 in the pruning of OPT models. While Proposition \ref{thm:alg} suggests that the theoretical complexity of Algorithm \ref{alg:localsearch} does not depend on $\hat t$ when $\hat t=\hat p$, we find that larger $\hat t$ values can expedite the algorithm in practice, due to benefits from vectorization. Consequently, for OPT models, practical choices of $\hat{k}=\hat{t}=10~\text{or}~100$ can offer an effective balance of performance and faster pruning times.

Next, we examine the impact of selecting a small $\hat p$ on the performance of our algorithm. In Algorithm \ref{alg:localsearch}, the sum of $p_i$ chosen in each iteration needs to be a fixed number. Thus, choosing for a small $\hat p$ for all iterations leads to a substantially larger number of iterations, potentially degrading performance. To address this, we consider the following two parameter settings:
\begin{enumerate}
    \item We set the number of iterations $T=p'/\hat t$ and set $t_i=p_i=\hat t$ for all iterations $i\in[T]$.
    \item We set the number of iterations $T=p'/\hat t+30$. For the first $p'/\hat t$ iterations, we set $t_i=p_i=\hat t$, and for the final 30 iterations, we set $t_i=\hat t$ and $p_i=0$.
\end{enumerate}
The first setting aligns with the parameters used in all our previous experiments. In the second setting, we initially follow the parameters of the first case, then switch to $\hat p=0$ for the last 30 iterations to implement a local swapping strategy. This involves exchanging elements within the set \( S \) with more impactful ones from outside \( S \) to achieve a set with a lower objective value, as depicted in the right part of Figure \ref{fig:step}. We refer to the first setting as "nested local search" and the second as "non-nested local search".

We assess the above two parameter settings in our ablation study. For ResNet20 and MobileNetV1, we set $\hat t=5$, and for ResNet50, $\hat t=10$. In the case of OPT models, we use $\hat t=2$ for pruning attention heads and $\hat t=100$ for reducing the intermediate dimension in fully connected layers. The accuracy results for pruning CNNs are presented in Table \ref{tab:cnn-non-nested-search}, and the perplexity results for pruning OPT models are presented in Table \ref{tab:llm-non-nested-search}.

We observe that \modelaname~with non-nested local search approach tends to yield better results in most scenarios, though the improvements are not always statistically significant. This finding suggests that in Algorithm \ref{alg:localsearch}, we may use a small $\hat p$ for non-nested local search to further refine the quality of solutions at the expense of more pruning time.

\begin{table}[!h]
\centering
    \begin{minipage}[b]{0.48\textwidth}\centering
        \centering
        \captionsetup{width=\textwidth}
\caption{Test accuracy for applying \modelaname~to prune CNNs (ResNet20, MobileNetV1 and ResNet50), with nested/non-nested local search. 
 For all models and parameters, we take ten runs and report the mean and standard error.}
\label{tab:cnn-non-nested-search}
\setlength{\tabcolsep}{5pt}
\resizebox{\textwidth}{!}{%
\begin{tabular}{c|c|c|c}
\toprule
Model & Speedup & \makecell{\modelaname\\ with \textit{nested} local search} & \makecell{\modelaname\\ with \textit{non-nested} local search} \\ \midrule
\multirow{6}{*}{ResNet20} 
&1.3x & \textbf{89.28} (±0.14) & 89.20 (±0.10)\\ 
&1.4x & \textbf{87.60} (±0.13) & 87.55 (±0.21)\\ 
&1.7x & \textbf{84.98} (±0.22) & 84.81 (±0.19)\\ 
&2.0x & 80.89 (±0.42) & \textbf{81.05} (±0.48)\\ 
&2.6x & 72.06 (±0.95) & \textbf{72.09} (±0.81)\\ 
&3.5x & 60.03 (±1.30) & \textbf{60.97} (±0.96) \\ \midrule
\multirow{6}{*}{MobileNetV1} 
&1.3x & 70.66 (±0.08) & \textbf{70.68} (±0.08)\\ 
&1.4x & 68.88 (±0.10) & \textbf{68.89} (±0.10)\\ 
&1.5x & 66.36 (±0.16) & \textbf{66.40} (±0.09)\\ 
&1.7x & 62.98 (±0.22) & \textbf{63.05} (±0.25)\\ 
&1.9x & 58.32 (±0.30) & \textbf{58.48} (±0.39)\\ 
&2.2x & 52.03 (±0.53) & \textbf{52.06} (±0.52)\\  \midrule
\multirow{7}{*}{ResNet50} 
&1.2x & 74.34 (±0.09) & \textbf{74.37} (±0.06)\\ 
&1.3x & 69.18 (±0.15) & \textbf{69.20} (±0.20)\\ 
&1.4x & \textbf{65.23} (±0.23) & 65.21 (±0.24)\\ 
&1.6x & 58.95 (±0.31) & \textbf{59.05} (±0.44)\\ 
&1.7x & 50.07 (±0.38) & \textbf{50.10} (±0.59)\\ 
&1.9x & 38.23 (±0.36) & \textbf{38.25} (±0.52)\\  \bottomrule
\end{tabular}%
}
    \end{minipage}
    \hfill
    \begin{minipage}[b]{0.48\textwidth}\centering
        \centering
        \captionsetup{width=\textwidth}
\caption{Perplexity performance on Wikitext for applying \modelaname~to prune OPT models (1.3B, 2.7B and 6.7B), with nested/non-nested local search. 
 For all models and parameters, we take ten runs and report the mean and standard error.}
\label{tab:llm-non-nested-search}
\setlength{\tabcolsep}{5pt}
\resizebox{\textwidth}{!}{%
\begin{tabular}{c|c|c|c}
\toprule
Model & Speedup & \begin{tabular}[c]{@{}c@{}}\modelaname\\ with \textit{nested} local search\end{tabular} & \begin{tabular}[c]{@{}c@{}}\modelaname\\ with \textit{non-nested} local search\end{tabular} \\ \midrule
\multirow{7}{*}{OPT-1.3B} & 1.2x & 15.50 ($\pm$0.09) & \textbf{15.38} ($\pm$0.06) \\
 & 1.3x & 17.40 ($\pm$ 0.11) & \textbf{17.30} ($\pm$0.10) \\
 & 1.4x & 20.76 ($\pm$0.20) & \textbf{20.62} ($\pm$0.18) \\
 & 1.7x & 26.83 ($\pm$0.32) & \textbf{26.55} ($\pm$0.35) \\
 & 2.0x & 39.96 ($\pm$0.65) & \textbf{39.38} ($\pm$0.60) \\
 & 2.6x & 76.51 ($\pm$1.97) & \textbf{73.66} ($\pm$1.86) \\
 & 3.3x & 160.9 ($\pm$6.12) & \textbf{156.7} ($\pm$4.57) \\ \midrule
\multirow{7}{*}{OPT-2.7B} & 1.2x & 13.30 ($\pm$0.08) & \textbf{13.17} ($\pm$0.07) \\
 & 1.3x & 15.17 ($\pm$0.13) & \textbf{15.05} ($\pm$0.13) \\
 & 1.4x & 17.43 ($\pm$0.14) & \textbf{17.33} ($\pm$0.16) \\
 & 1.7x & 21.88 ($\pm$0.25) & \textbf{21.68} ($\pm$0.26) \\
 & 2.0x & 30.58 ($\pm$0.33) & \textbf{30.24} ($\pm$0.33) \\
 & 2.4x & 54.90 ($\pm$1.04) & \textbf{53.11} ($\pm$0.82) \\
 & 3.0x & 107.0 ($\pm$2.35) & \textbf{105.3} ($\pm$2.77) \\ \midrule
\multirow{7}{*}{OPT-6.7B} & 1.1x & 11.55 ($\pm$0.03) & \textbf{11.36} ($\pm$0.03) \\
 & 1.2x & 12.98 ($\pm$0.08) & \textbf{12.79} ($\pm$0.08) \\
 & 1.5x & 15.00 ($\pm$0.10) & \textbf{14.94} ($\pm$0.13) \\
 & 1.7x & 18.97 ($\pm$0.23) & \textbf{18.92} ($\pm$0.22) \\
 & 2.0x & 26.01 ($\pm$0.36) & \textbf{26.05} ($\pm$0.36) \\
 & 2.6x & 47.55 ($\pm$0.72) & \textbf{46.63} ($\pm$0.76) \\
 & 3.3x & 106.9 ($\pm$3.88) & \textbf{100.8} ($\pm$3.01) \\ \bottomrule
\end{tabular}%
}
    \end{minipage}%
\end{table}

\end{document}